\documentclass{article}

\usepackage{iclr2027_conference,times}
\iclrfinalcopy
\usepackage{caption}
\usepackage[utf8]{inputenc}
\usepackage[T1]{fontenc}
\usepackage{hyperref}
\usepackage{url}
\usepackage{subcaption}
\usepackage{booktabs}
\usepackage{amsmath,amssymb,amsthm}
\usepackage{amsfonts}
\usepackage{nicefrac}
\usepackage{microtype}
\usepackage{xcolor}
\usepackage{graphicx}
\usepackage{enumitem}
\usepackage{bbm}
\usepackage[ruled,vlined]{algorithm2e}
\usepackage{algorithmic}
\usepackage{threeparttable}
\usepackage{array}
\usepackage[capitalise,noabbrev]{cleveref}
\numberwithin{equation}{section}
\newtheorem{theorem}{Theorem}[section]
\newtheorem{proposition}[theorem]{Proposition}
\newtheorem{corollary}[theorem]{Corollary}
\newtheorem{lemma}[theorem]{Lemma}

\theoremstyle{remark}

\title{Rethinking Optimization Methods for Hierarchical Feature Selection}

\author{\textbf{Xinyuan Song}$^{1}$ \quad
    \textbf{Zekun Cai}$^{2}$ \quad
    \textbf{Liang Zhao}$^{1}$ \\
    $^{1}$Emory University, Atlanta, GA, USA \quad
    $^{2}$The University of Tokyo, Tokyo, Japan \\
    \texttt{\{xinyuan.song,liang.zhao\}@emory.edu, caizekun@csis.u-tokyo.ac.jp} \\
}

\begin{document}

\maketitle

\begin{abstract}
Rare-event forecasting often depends on interpretable conjunctions: a single keyword, location, or temporal cue may be weak, while their joint occurrence can identify a meaningful precursor. Hierarchical-Task Numerical conjunctive feature Learning (HTNL) selects Numerical Conjunctive Features (NCFs) through a structured sparsity objective. However, the resulting optimization problem is challenging, making optimization design critical to its practical performance. We give a unified diagnosis of four HTNL optimization families: original majorization-minimization (MM), variational quadratic form (VQF), alternating direction method of multipliers (ADMM), and weighted group lasso (WGL). Our analysis distinguishes three key factors that jointly determine HTNL optimization behavior. First, VQF and MM are algebraically equivalent under matched variational updates and optimize a squared-penalty formulation. Second, ADMM optimizes the unsquared objective through consensus splitting; we replace its truncated FISTA proximal solver with a scalar bisection update (ADMM-Bisect) for more accurate treatment of near-threshold blocks. Third, WGL targets the same unsquared objective, while all methods face rapidly decaying active-set admission scores. Across two synthetic datasets and 11 civil-unrest and flu-surveillance tasks, VQF achieves the highest mean AUC with the most compact non-collapsing supports. ADMM-Bisect reduces measured runtime, while calibrated WGL avoids the empty-support failures of its original scale. Synthetic results further show how proximal accuracy, pruning, and search budget affect candidate recovery, especially for longer conjunctions. Code will be made publicly available upon publication.
\end{abstract}

% =============================================================================

\section{Introduction}

\label{sec:intro}

Forecasting rare localized phenomena such as civil unrest~\citep{Compton2014,Ramakrishnan2014,Cadena2015,Goode2015}, disease outbreaks~\citep{Achrekar2011,Lamb2013}, and disaster responses~\citep{Sakaki2010} requires models that can identify weak but meaningful precursor patterns in noisy spatiotemporal streams~\citep{Zhao2015stef,Zhao2016mresolution,Ning2018staple}. In these domains, a single keyword, location, or temporal cue is rarely decisive. Predictive signal often appears only through conjunctions~\citep{Zhao2019htnl}: a protest-related phrase in a particular country during a narrow time window, or a flu-related phrase co-occurring with a regional surveillance pattern. Numerical Conjunctive Features (NCFs), defined as products of primitive co-occurrence signals, give a direct and interpretable representation of such multi-factor precursors for spatial event forecasting~\citep{Zhao2019htnl}.

Hierarchical-Task Numerical conjunctive feature Learning (HTNL)~\citep{Zhao2019htnl} addresses this problem by selecting predictive conjunctions from an exponentially large NCF space without explicitly enumerating all candidates. Its hierarchical sparsity structure provides two key benefits: it favors compact, interpretable sets of precursor conjunctions and simultaneously identifies the geographic groups in which those conjunctions are informative. This makes HTNL particularly suitable for rare-event forecasting. The difficulty is optimization. HTNL combines an outer hierarchy over NCFs with nested geographic group norms, producing strongly coupled structured-sparsity penalties for which standard lasso or group-lasso updates~\citep{Tibshirani1996,Yuan2006,Jacob2009,Jenatton2011,Bach2012} cannot be applied directly. Moreover, because the exponentially large feature space must be explored through active-set expansion, performance depends not only on solving the current optimization problem accurately but also on whether useful unseen conjunctions can enter the candidate set. 

% This paper argues that HTNL should be understood as an optimization system, not merely as a regularizer. Existing implementations expose several solver choices: majorize-minimize (MM), variational quadratic form (VQF), alternating direction method of multipliers (ADMM), and weighted group lasso (WGL) continuation. These solvers attack the same nested objective from different angles: MM and VQF use variational quadratic reformulations, ADMM uses a consensus split for the nonsmooth objective~\citep{Boyd2011,EcksteinBertsekas1992,Parikh2014}, and WGL reduces the hierarchy to a weighted one-level regularization path related to classical path algorithms~\citep{Friedman2010}. The resulting empirical differences can come from very different sources. Some are algebraic equivalences, some are numerical errors in an inner proximal step, and some are structural limits of the outer active-set search.

Common optimization strategies for structured sparsity include majorization--minimization, variational reformulation, operator-splitting methods such as ADMM, and weighted group-lasso continuation~\citep{Hunter2004,Boyd2011,EcksteinBertsekas1992,Parikh2014,Friedman2010}. For HTNL, we therefore compare four corresponding solvers: MM (introduced in~\citep{Zhao2019htnl}), VQF, ADMM, and WGL continuation. MM and VQF use quadratic surrogate updates, ADMM separates the smooth loss from the nonsmooth hierarchical penalty through consensus splitting, and WGL solves a sequence of weighted group-lasso problems along a regularization path.

We diagnose these solvers at three levels: the objective formulation, the inner nonsmooth update, and the outer active-set search. At the formulation level, MM and VQF are not distinct optimization methods: under matched auxiliary variables, they yield the same weighted-ridge updates, following the broader connection between variational reformulation and majorization--minimization~\citep{Hunter2004}. At the solver level, ADMM's truncated FISTA proximal step~\citep{Beck2009} can leave near-threshold blocks numerically nonzero. We replace this approximation with a scalar bisection solver (ADMM-Bisect), which computes the proximal update to numerical precision. WGL exhibits a different issue: its continuation path is highly sensitive to regularization scale and can collapse to an empty support when the initial penalty is too large. Beyond these solver-specific effects, all methods must discover unseen NCFs from an exponentially large candidate space, connecting HTNL to feature screening methods for large structured feature spaces~\citep{Tibshirani2012strong}.

We evaluate these mechanisms on two synthetic datasets and 11 real-world rare-event tasks spanning Latin American civil unrest~\citep{Compton2014,Ramakrishnan2014} and US flu surveillance~\citep{Achrekar2011,Lamb2013}. The synthetic experiments isolate proximal accuracy, pruning, and search-budget effects, while the real-world tasks evaluate prediction, sparsity, and runtime. Non-collapsing HTNL variants outperform a class-weighted logistic baseline, with VQF achieving the highest mean AUC and the most compact supports. ADMM-Bisect removes proximal residuals with lower measured runtime, while regularization calibration prevents WGL from collapsing to empty supports.

This paper makes three main contributions:
\begin{enumerate}[leftmargin=*,nosep]
  \item \textbf{Unified optimization diagnosis.}
	  We study four HTNL optimization families-MM, VQF, ADMM, and WGL continuation-on rare-event NCF selection tasks, and show how they split into equivalent, sparse-selector, over-selector, and collapse regimes.

  \item \textbf{Exact ADMM proximal update.}
  We introduce an exact bisection proximal operator for the ADMM subproblem and combine it with KKT-pruning, yielding a more reliable and faster ADMM variant while preserving the original HTNL objective.

  \item \textbf{Structural active-set bound.}
  We prove that the active-set KKT expansion score decays rapidly with NCF length, giving a structural explanation for the observed performance ceiling and showing that breaking this ceiling requires redesigning the active-set expansion rule.
\end{enumerate}

\section{Background}

\label{sec:background}

\subsection{HTNL Model}

Let $\mathbf{X}\in\mathbb{R}^{T\times S\times P}$ be the input tensor, where $T$ is the number of time steps, $S$ is the number of geographic locations, and $P$ is the number of primitive features. We write $X_{s,t}\in\mathbb{R}^{P}$ for the feature vector at location $s$ and time $t$, and $Y_{s,t}$ for the corresponding binary event label.

A Numerical Conjunctive Feature (NCF) is a product of primitive features. For an NCF $v=\{j_1,\ldots,j_k\}\subseteq[P]$, its value is
\begin{equation} \phi_v(X_{s,t})=\prod_{j\in v}x_j(s,t). \label{eq:ncf_definition} \end{equation}
Let $\mathcal{V}$ be the candidate NCF set and let $D(v)$ and $A(v)$ denote the descendant and ancestor sets of $v$ in the NCF lattice.

Hierarchical-Task Numerical conjunctive feature Learning (HTNL) learns a location-specific weight $W_{s,v}$ for each NCF $v$. The prediction score is
\begin{equation} f_{s,t}(W)=\sum_{v\in\mathcal{V}}W_{s,v}\phi_v(X_{s,t})+b_s. \label{eq:htnl_prediction} \end{equation}
The empirical loss is
\begin{equation} \mathcal{L}(W)=\sum_{s=1}^{S}\sum_{t=1}^{T}C_{s,t}\ell(Y_{s,t},f_{s,t}(W)), \label{eq:empirical_risk} \end{equation}
where $\ell$ is a convex classification loss and $C_{s,t}$ is a sample weight.

\subsection{Bi-Level Regularizer}

HTNL uses a bi-level sparsity penalty to couple NCF selection and geographic structure. Let $\mathcal{G}$ be the set of geographic groups. For each NCF $u$, define
\begin{equation} r_u(W)=\sum_{g\in\mathcal{G}}\left\|\{W_{s,u}\}_{s\in g}\right\|_2. \label{eq:geo_group_norm} \end{equation}
For each ancestor $v$, the local hierarchical penalty is
\begin{equation} \Omega_v(W)=d_v\left(\sum_{u\in D(v)}r_u(W)^p\right)^{1/p},\qquad 1<p\le 2, \label{eq:local_htnl_regularizer} \end{equation}
where $d_v=a^{|v|}$ is a depth-dependent penalty weight. The full HTNL regularizer is
\begin{equation} \Omega(W)=\sum_{v\in\mathcal{V}}d_v\left(\sum_{u\in D(v)}\left(\sum_{g\in\mathcal{G}}\left\|\{W_{s,u}\}_{s\in g}\right\|_2\right)^p\right)^{1/p}. \label{eq:omega-two-level} \end{equation}
The learning objective is
\begin{equation} \min_W \mathcal{L}(W)+\lambda\Omega(W). \label{eq:htnl_objective} \end{equation}

\subsection{Active-Set Framework}

Because the NCF lattice is combinatorial, HTNL optimizes over a working set $A^{(m)}\subseteq\mathcal{V}$ at outer iteration $m$:
\begin{equation} \min_{W_{A^{(m)}}}\mathcal{L}(W_{A^{(m)}})+\lambda\Omega(W_{A^{(m)}}). \label{eq:restricted_htnl_objective} \end{equation}
After solving the restricted problem, inactive candidates are scored by their KKT violation:
\begin{equation} \mathrm{score}(v)=\frac{1}{d_v}\left\|\sum_{s=1}^{S}\sum_{t=1}^{T}\nabla_W\mathcal{L}(W^*)\phi_v(X_{s,t})\right\|_2. \label{eq:kkt_score_background} \end{equation}
Candidates with $\mathrm{score}(v)>\tau_{\mathrm{KKT}}$ are added:
\begin{equation} A^{(m+1)}=A^{(m)}\cup\{v\notin A^{(m)}:\mathrm{score}(v)>\tau_{\mathrm{KKT}}\}. \label{eq:active_set_update} \end{equation}
The loop stops when no new candidate violates the threshold or when the maximum number of outer rounds is reached. All solvers studied in this paper share this active-set framework and differ only in how they solve the restricted inner problem.

\section{Methodology}\label{sec:method}

\begin{figure}[!ht]
  \centering
  \includegraphics[width=\linewidth]{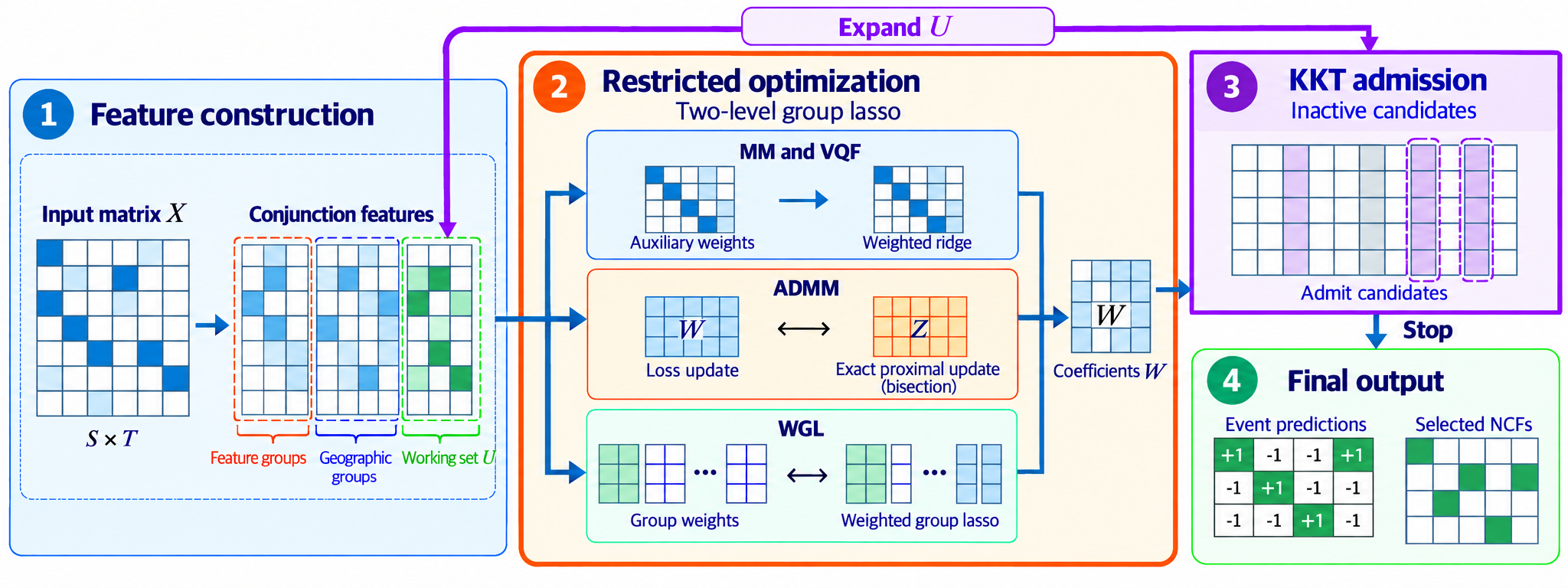}

\caption{\textbf{Main pipeline of HTNL optimization and NCF selection.}
HTNL constructs conjunction features from $X$ and optimizes the restricted two-level group-lasso problem using MM--VQF, ADMM, or WGL. The resulting coefficients score unseen NCFs by the KKT criterion, admitted candidates expand the working set $U$, and the process repeats until convergence, producing event predictions and selected NCFs.}
  \label{fig:pipeline}

\end{figure}

Figure~\ref{fig:pipeline} summarizes the optimization pipeline. Starting from the HTNL objective in~\eqref{eq:htnl_objective}, we solve the restricted problem in~\eqref{eq:restricted_htnl_objective} using three approaches: VQF converts the two-level regularizer in~\eqref{eq:omega-two-level} into weighted ridge updates, ADMM retains the nonsmooth hierarchy through consensus splitting, and WGL reduces it to a weighted group-lasso path. All three feed into the same outer update in~\eqref{eq:active_set_update}.

\subsection{Variational Quadratic Form}\label{sec:vqf_method}

We first derive a variational quadratic form for the HTNL regularizer. The goal is to transform the nested nonsmooth penalty in~\eqref{eq:omega-two-level} into a weighted quadratic penalty after introducing auxiliary variables. This gives a majorize-minimize-style update: for fixed auxiliary variables, the inner optimization becomes a weighted ridge problem in $W$. This is the same weighted-ridge reduction used by the original MM solver in~\citep{Zhao2019htnl}, and Proposition~\ref{prop:equiv} shows that the two formulations produce identical $W$-updates under the same auxiliary-weight update rule.

\begin{lemma}[Variational quadratic form of a two-level group lasso]\label{lemma:1}
Let $W=\{W_{j,u}\}$, where $u\in\mathcal V$ indexes conjunctive features and $j\in g$ indexes locations in geographical group $g\in\{1,\ldots,G\}$. For each $v\in\mathcal V$, let $D(v)\subseteq\mathcal V$ be the descendant set of $v$, and let $d_v>0$. Recall $\Omega(W)$ has the form Equation~\eqref{eq:omega-two-level} with $1<p\le 2$. Let
\begin{equation} \Theta_{m,r}=\left\{x\in\mathbb R_+^m:\sum_{i=1}^m x_i^r\le 1\right\}. \label{eq:theta-set} \end{equation}
Then $\Omega(W)^2$ admits the equivalent quadratic representation
\begin{equation} \Omega(W)^2=\min_{\gamma,\lambda,\mu}\sum_{v\in\mathcal V}\frac{d_v^2}{\gamma_v}\sum_{u\in D(v)}\frac{1}{\lambda_{v,u}}\sum_{g=1}^G\frac{1}{\mu_{u,v,g}}\sum_{j\in g}\|W_{j,u}\|_2^2, \label{eq:omega-quadratic-representation} \end{equation}
where $\gamma\in\Theta_{|\mathcal V|,1}$,$ \lambda_v\in\Theta_{|D(v)|,\bar p}$,$\mu_{u,v}\in\Theta_{G,1}$, $\bar p=\frac{p}{2-p}$. Thus, for fixed $\gamma,\lambda,\mu$, the regularizer is a weighted ridge penalty in $W$.
\end{lemma}
The proof is deferred to Appendix~\ref{app:lemma1}.

\begin{proposition}[Equivalence of VQF and MM]
\label{prop:equiv}
Let $\Omega(W)$ be the HTNL regularizer in~\eqref{eq:omega-two-level}, and let its variational quadratic representation be given by~\eqref{eq:omega-quadratic-representation}. Suppose VQF and MM start from the same initialization $W^{(0)}$ and update the auxiliary variables $(\gamma,\lambda,\mu)$ by the same exact minimization of the variational representation at each iterate. Then, for every iteration $t\ge 0$, the two methods solve the same weighted-ridge subproblem and produce identical iterates:
\begin{equation} W_{\mathrm{VQF}}^{(t)}=W_{\mathrm{MM}}^{(t)}. \label{eq:vqf_mm_equiv} \end{equation}
\end{proposition}
The proof is deferred to Appendix~\ref{app:prop1}; algebraically, the two methods yield identical auxiliary weights and therefore the same weighted-ridge update at every iteration.

\subsection{Direct ADMM Optimization}\label{sec:admm_method}

We next consider ADMM, which directly optimizes the nonsmooth two-level group-lasso objective rather than replacing it with a quadratic surrogate. On a fixed candidate set $U$, ADMM introduces consensus copies $Z_{v,u,g}=P_{u,g}W_U$ to separate the smooth loss from the hierarchical penalty~\citep{EcksteinBertsekas1992,Boyd2011,Parikh2014}. The $W$-update is therefore smooth, while the original nested regularizer is retained in the $Z$-update.

\begin{lemma}[Consensus decomposition of the HTNL objective]\label{lemma:admm_consensus}
Let $D_U(v)=D(v)\cap U$ and define the local two-level penalty $\Omega_v(Z_v)=\left(\sum_{u\in D_U(v)}\left(\sum_{g\in\mathcal G}\|Z_{v,u,g}\|_2\right)^p\right)^{1/p}$. Introducing $Z_{v,u,g}=P_{u,g}W_U$ gives the ADMM updates
\begin{equation} W_U^{k+1}=\arg\min_{W_U}\mathcal L(W_U)+\frac{\rho}{2}\sum_{v\in U}\sum_{u\in D_U(v)}\sum_{g\in\mathcal G}\|P_{u,g}W_U-Z_{v,u,g}^k+Y_{v,u,g}^k\|_2^2. \label{eq:main-admm-w} \end{equation}
\begin{equation} Z_v^{k+1}=\operatorname{prox}_{(\lambda d_v/\rho)\Omega_v}(Q_v^{k+1}),\qquad Q_{v,u,g}^{k+1}=P_{u,g}W_U^{k+1}+Y_{v,u,g}^k. \label{eq:main-admm-z} \end{equation}
Thus, ADMM isolates the nonsmooth hierarchy in a local proximal problem while leaving the parameter update smooth.
\end{lemma}
The complete consensus formulation, dual update, and stopping conditions are given in Appendix~\ref{sec:direct_admm}.

The update in~\eqref{eq:main-admm-z} is exact only when its proximal problem is solved exactly. In practice, however, the published ADMM implementation~\citep{Zhao2019htnl} approximates this update using a fixed number of FISTA iterations~\citep{Beck2009}. We next quantify the resulting finite-step error and show why it can affect NCF support recovery near the zero/nonzero boundary.

\subsection{Finite-Step Proximal Error in ADMM}
\label{sec:prop2}

Although~\eqref{eq:main-admm-z} defines the exact proximal update, the published implementation approximates it with a fixed number of FISTA iterations~\citep{Beck2009}. To quantify the resulting error, consider the consensus form
\begin{equation} \mathcal{L}_{\rho}(W,Z,\Lambda)=\mathcal{L}(W)+\lambda\Omega^2(Z)+\frac{\rho}{2}\|W-Z+\Lambda\|_2^2. \label{eq:augmented_lagrangian} \end{equation}
For fixed $W$ and $\Lambda$, the exact $Z$-update is
\begin{equation} Z^*=\operatorname{prox}_{\lambda\Omega^2/\rho}(Q)=\arg\min_Z\left\{\lambda\Omega^2(Z)+\frac{\rho}{2}\|Z-Q\|_2^2\right\},\qquad Q=W+\Lambda. \label{eq:z_prox} \end{equation}
Let $Z^{(K)}$ be the output after $K$ FISTA iterations. Define the proximal objective and its finite-step error as
\begin{equation} \mathcal{F}_{\rho}(Z)=\lambda\Omega^2(Z)+\frac{\rho}{2}\|Z-Q\|_2^2,\qquad \varepsilon_K=\mathcal{F}_{\rho}(Z^{(K)})-\mathcal{F}_{\rho}(Z^*). \label{eq:inner_objective_error} \end{equation}
The following proposition translates this objective error into a bound on the proximal iterate error.

\begin{proposition}[Finite-step proximal error in ADMM]
\label{prop:admm_residual}
Since $\mathcal{F}_{\rho}$ in~\eqref{eq:inner_objective_error} is $\rho$-strongly convex,
\begin{equation} \|Z^{(K)}-Z^*\|_2\le \sqrt{\frac{2\varepsilon_K}{\rho}}. \label{eq:proximal_error_bound} \end{equation}
If the inner method satisfies $\varepsilon_K=O(K^{-2})$, then
\begin{equation} \|Z^{(K)}-Z^*\|_2=O\left(\frac{1}{\sqrt{\rho}K}\right). \label{eq:admm_finite_step_rate} \end{equation}
\end{proposition}

Consequently, Proposition~\ref{prop:admm_residual} shows that a fixed-budget inner solve can leave a nonzero discrepancy $\|Z^{(K)}-Z^*\|_2$, so a block sufficiently close to the zero/nonzero boundary may remain numerically nonzero even when its exact proximal solution is zero. The proof is deferred to Appendix~\ref{app:prop2}. The effect of this finite-step error on support recovery therefore depends on the zero/nonzero boundary of the exact proximal problem.

\subsection{Zero-Block Recovery in ADMM}\label{sec:kkt_margin}

We answer this question through the KKT condition of the exact proximal problem in~\eqref{eq:z_prox}. In particular, the distance of $Z^*$ from the zero/nonzero boundary determines whether the approximation error bounded in Proposition~\ref{prop:admm_residual} can change a block's support status. This connects finite-step proximal accuracy directly to exact zero-block recovery.

\begin{lemma}[Zero-Block Recovery in ADMM]\label{lemma:kkt_margin}
For the ADMM proximal problem with exact solution $Z^*$ and proximal input $Q$, the KKT condition is
\begin{equation} 0\in\rho(Z^*-Q)+\lambda\partial\Omega^2(Z^*). \label{eq:main-prox-kkt} \end{equation}
For a block $u$, $Z_u^*=0$ only when $\rho Q_u$ lies in the corresponding zero-block subgradient set $\lambda\partial_u\Omega^2(0;Z^*_{-u})$. Hence, if the finite-step proximal error is smaller than the distance to this zero/nonzero boundary, the approximate and exact proximal updates recover the same support up to numerical tolerance. A useful local scale for this regime is
\begin{equation} \rho_c^{\mathrm{proxy}}=\frac{d_v}{\|f_v(W)^{p-1}\|_{\infty}},\qquad f_v(W)=\left(\sum_{u\in D(v)}\|W_u\|_2^p\right)^{1/p}. \label{eq:main-rho-proxy} \end{equation}
\end{lemma}
This result explains why improving proximal accuracy can remove numerical over-selection without changing the underlying HTNL objective. The proof is provided in Appendix~\ref{sec:kkt_margin_hard_zero}.

\subsection{ADMM-Bisect: Exact Proximal Update}\label{sec:admm_bisect_contribution}

Proposition~\ref{prop:admm_residual} shows that a fixed-budget FISTA solve can leave a nonzero proximal residual, while Lemma~\ref{lemma:kkt_margin} shows that such an error can change the zero/nonzero decision for blocks near the KKT boundary. We therefore replace the fixed-budget FISTA approximation in the ADMM $Z$-update with a bisection solver~\citep{burden2015numerical} for the local proximal equations. By solving these equations to numerical tolerance, ADMM-Bisect removes the truncation residual that can cross the KKT boundary and thereby prevents solver-induced retention of blocks whose exact proximal solution is zero.

For the single-geographic-group, scalar-coefficient adapter, let $q=p/(p-1)$ and $c=\lambda d_v/\rho$. The exact local proximal solution satisfies $Z_v^*=0$ when $\|Q_v\|_q\le c$. Otherwise, when $\|Q_v\|_q>c$, each coordinate has the form
\begin{equation} Z_{v,u}^*=\operatorname{sign}(Q_{v,u})r_u,\qquad r_u+\eta r_u^{p-1}=|Q_{v,u}|,\qquad r_u\ge0. \label{eq:bisect-scalar-equation} \end{equation}
The coordinates are coupled through
\begin{equation} \eta=\frac{c}{\left(\sum_u r_u^p\right)^{(p-1)/p}}. \label{eq:bisect-eta-consistency} \end{equation}
For fixed $\eta>0$, the scalar equation in~\eqref{eq:bisect-scalar-equation} is monotone and has a unique root in $[0,|Q_{v,u}|]$, which is solved by bisection. We update $\eta$ using~\eqref{eq:bisect-eta-consistency} until the local norm is consistent. Since bisection requires $O(\log\varepsilon^{-1})$ iterations to reach tolerance $\varepsilon$, solving $|D_U(v)|$ scalar equations over $B_v$ norm-consistency updates costs $O(B_v|D_U(v)|\log\varepsilon^{-1})$ for each local hierarchical block.

\subsection{Weighted Group-Lasso Reduction}\label{sec:wgl_method}

An alternative is to reduce the two-level hierarchy to a sequence of weighted one-level group-lasso problems. Using the dual representation of the outer $\ell_p$ norm, WGL fixes dual weights for the hierarchical level, solves the resulting weighted group lasso, and then updates those weights in closed form. This connects the HTNL hierarchy to standard group-lasso and regularization-path optimization~\citep{Yuan2006,Friedman2010}.

\begin{lemma}[Weighted one-level reduction]\label{lemma:wgl_reduction}
Let $r_{v,u}(W_U)=\sum_{g\in\mathcal G}\|P_{u,g}W_U\|_2$ and let $q$ satisfy $1/p+1/q=1$. Then the local hierarchical penalty satisfies
\begin{equation} \left(\sum_{u\in D_U(v)}r_{v,u}(W_U)^p\right)^{1/p}=\max_{\alpha_v\ge0,\ \|\alpha_v\|_q\le1}\sum_{u\in D_U(v)}\alpha_{v,u}r_{v,u}(W_U). \label{eq:main-wgl-dual} \end{equation}
For fixed $\alpha$, the restricted HTNL problem reduces to
\begin{equation} \min_{W_U}\mathcal L(W_U)+\sum_{u\in U}\sum_{g\in\mathcal G}c_{u,g}(\alpha)\|P_{u,g}W_U\|_2,\qquad c_{u,g}(\alpha)=\lambda\sum_{v\in A_U(u)}d_v\alpha_{v,u}. \label{eq:main-wgl-reduction} \end{equation}
Thus, fixing the outer dual weights converts the two-level penalty into a weighted one-level group lasso.
\end{lemma}
WGL alternates the weighted group-lasso solve with closed-form updates of $\alpha$ and applies this procedure along a continuation path. The full reduction and algorithm are given in Appendix~\ref{sec:wgl_reduction}.

\subsection{Active-Set Accessibility Limit}
\label{sec:prop3}

The preceding methods differ in how they solve the restricted HTNL problem, but all share the active-set expansion rule in~\eqref{eq:active_set_update}. Thus, improving the inner optimization does not guarantee that a predictive NCF can enter the working set. By~\eqref{eq:kkt_score_background}, an inactive NCF is admitted only when its KKT score exceeds $\tau_{\mathrm{KKT}}$. We characterize this shared accessibility constraint by showing that the admission score decays rapidly with conjunction length.

Let $v\subseteq[P]$ be an NCF of length $k=|v|$, with $\phi_v(X_{s,t})=\prod_{j\in v}x_j(s,t)$. Assume that $x_j(s,t)\in\{0,1\}$ are independent Bernoulli variables with $\Pr[x_j(s,t)=1]=p_j\le p_{\max}<1$. Let $n_{\mathrm{eff}}=|S||T|$ and assume that the loss gradient is uniformly bounded:
\begin{equation} \left\|\nabla_W\mathcal{L}(W^*)_{s,t}\right\|_2\le G \qquad \text{for all } (s,t). \label{eq:gradient_bound_prop3} \end{equation}
Under these assumptions, the KKT admission score in~\eqref{eq:kkt_score_background} admits the following length-dependent bound.

\begin{proposition}[Exponential decay of active-set KKT scores]
\label{prop:ceiling}
Under the assumptions above, the expected KKT score of a length-$k$ NCF satisfies
\begin{equation} \mathbb{E}[\mathrm{score}(v)]\le G n_{\mathrm{eff}}\left(\frac{p_{\max}}{2}\right)^k. \label{eq:kkt_decay_rigorous} \end{equation}
Consequently, for any threshold $\tau_{\mathrm{KKT}}>0$, if
$k\ge k^*:=\left\lceil\frac{\log(G n_{\mathrm{eff}}/\tau_{\mathrm{KKT}})}{\log(2/p_{\max})}\right\rceil$, 
then $\mathbb{E}[\mathrm{score}(v)]\le \tau_{\mathrm{KKT}}$. Moreover, for any $\epsilon>0$,
\begin{equation} \Pr\left[\mathrm{score}(v)\ge \tau_{\mathrm{KKT}}+\epsilon\right]\le \frac{G n_{\mathrm{eff}}\left(p_{\max}/2\right)^k}{\tau_{\mathrm{KKT}}+\epsilon}. \label{eq:kkt_markov_bound} \end{equation}
\end{proposition}

Equations~\eqref{eq:kkt_decay_rigorous}--\eqref{eq:kkt_markov_bound} show that both the expected KKT score and the probability of clearing the threshold decay exponentially with NCF length. Thus, Proposition~\ref{prop:ceiling} establishes an accessibility bias against long conjunctions shared by all solvers, explaining their saturation when useful NCFs cannot enter the active set. The proof is deferred to Appendix~\ref{app:prop3}.

\begin{corollary}[Active-set accessibility bound]
\label{cor:active_set_accessibility}
Let $U^*$ be the true NCF support, and define the accessible true support as
\begin{equation} U^*_{\mathrm{acc}}:=\left\{u\in U^*: |u|\le D \text{ and } \mathbb{E}[\mathrm{score}(u)]>\tau_{\mathrm{KKT}}\right\}, \label{eq:accessible_support_def} \end{equation}
where $D$ is the maximum NCF length allowed by the active-set search. Any optimization using this active-set rule can recover true NCFs only from $U^*_{\mathrm{acc}}$ in expectation. Hence the expected true-support recall is bounded by
\begin{equation} \mathbb{E}\left[\mathrm{Recall}(\hat{S};U^*)\right]\le \frac{|U^*_{\mathrm{acc}}|}{|U^*|}. \label{eq:accessible_recall_bound} \end{equation}
\end{corollary}

Together, Proposition~\ref{prop:ceiling} and Corollary~\ref{cor:active_set_accessibility} show that exponential KKT-score decay restricts recovery to the accessible support $U^*_{\mathrm{acc}}$, imposing the recall ceiling in~\eqref{eq:accessible_recall_bound}. Thus, improving the inner optimization alone cannot recover predictive NCFs excluded by the shared active-set rule. The proof is deferred to Appendix~\ref{app:active_set_accessibility}.

\section{Controlled Optimization Method Diagnostics}

\label{sec:synthetic_diagnostics}

Before turning to real data, we use two controlled synthetic datasets with paired seeds to isolate the mechanisms predicted by our analysis. These diagnostics distinguish algebraic equivalence, finite-step numerical error, and regularization-path effects under controlled conditions.

Table~\ref{tab:main_results} validates Proposition~\ref{prop:equiv}, with MM and VQF remaining bit-exact under matched auxiliary updates, and tests Proposition~\ref{prop:admm_residual}, with exact bisection yielding the only positive F1 improvement that survives Holm--Bonferroni correction. The WGL variants further diagnose regularization-path sensitivity, showing that adaptive penalties do not reliably prevent collapse.

\begin{table}[!ht]
\centering
\caption{NCF support-selection F1 on two controlled synthetic datasets. Results are mean $\pm$ standard deviation over 20 paired random seeds at active-set thresholds $\tau=0.20$ and $0.40$.}

\label{tab:main_results}
\setlength{\tabcolsep}{5pt}
\resizebox{\linewidth}{!}{%
\begin{tabular}{lcccc}
\toprule
& \multicolumn{2}{c}{\textbf{Dataset 1}}
& \multicolumn{2}{c}{\textbf{Dataset 2}} \\
\cmidrule(lr){2-3}\cmidrule(lr){4-5}
\textbf{Method} & $\tau=0.20$ & $\tau=0.40$ & $\tau=0.20$ & $\tau=0.40$ \\
\midrule
HTNL-MM      & $0.580\pm0.201$ & $0.592\pm0.253$ & $0.581\pm0.188$ & $0.590\pm0.246$ \\
HTNL-VQF           & $0.580\pm0.201$ & $0.592\pm0.253$ & $0.581\pm0.188$ & $0.590\pm0.246$ \\
HTNL-ADMM (FISTA, $\rho=10$) & $0.468\pm0.133$ & $\mathbf{0.674\pm0.187}$ & $0.457\pm0.147$ & $\mathbf{0.674\pm0.211}$ \\
HTNL-WGL ($\lambda=100$)     & $0.463\pm0.336$ & $0.463\pm0.336$ & $0.360\pm0.306$ & $0.360\pm0.306$ \\
\midrule
\multicolumn{5}{l}{\textit{Best-of-class variants (post-hoc sweeps)}} \\
ADMM-Bisect ($\rho=1$, exact prox) & $0.488\pm0.128$ & $0.685\pm0.192$ & $0.468\pm0.137$ & $0.687\pm0.213$ \\
WGL-Cont (high-to-low, $500{\to}50$)       & $0.602\pm0.250$ & $0.549\pm0.273$ & $0.607\pm0.231$ & $0.567\pm0.273$ \\
WGL-Adaptive (0.2--5.0)                    & $0.403\pm0.195$ & $0.403\pm0.195$ & $0.383\pm0.214$ & $0.383\pm0.214$ \\
\bottomrule
\end{tabular}
}

\end{table}

Figures~\ref{fig:admm_paired} and~\ref{fig:kkt_speedup} isolate the ADMM mechanism. Consistent with Proposition~\ref{prop:admm_residual}, exact bisection never underperforms finite-step FISTA across paired seeds in the significant comparison. Combined with KKT pruning, it preserves F1 while reducing runtime by excluding low-violation blocks before the inner solve.

\section{Experiments}
\label{sec:experiments}

\subsection{Experimental Setup}
\label{sec:setup}

We evaluate the HTNL optimization methods on \textbf{11 real-world binary classification datasets}~\citep{zhao2014unsupervised,compton2014publicly}: 10 country-level Latin American Twitter civil-unrest corpora and one US flu-surveillance dataset. The civil-unrest datasets contain $V{=}100$ primitive feature groups and exhibit severe class imbalance, with positive rates of $0.5$--$1.5\%$. The flu dataset contains $V{=}182$ feature groups, $n_{\mathrm{train}}{=}5{,}096$, and a positive rate of $7.4\%$. These datasets span severely imbalanced event forecasting and a comparatively less imbalanced public-health prediction setting.

We use held-out AUC as the primary evaluation metric because it provides a threshold-independent assessment of ranking performance~\citep{Hanley1982,Fawcett2006}. Given the severe class imbalance, we additionally report precision--recall-based metrics where appropriate, which complement ROC-based evaluation in highly skewed settings~\citep{DavisGoadrich2006}. All HTNL variants use the same class-weighted Huber-hinge objective~\citep{Huber1964} without resampling. Statistical comparisons against Logistic Regression use paired $t$-tests across the 11 datasets, with Bonferroni correction~\citep{Bonferroni1936} over four pre-specified HTNL comparisons.

Unless stated otherwise, all methods use the same active-set configuration. WGL follows a four-step continuation schedule from $\lambda_{\mathrm{init}}{=}500$ to $\lambda_{\mathrm{target}}{=}50$; we additionally evaluate a scale-corrected $\lambda{=}1$ variant motivated by the class-imbalance analysis in Section~\ref{app:wgl_sensitivity}.

\subsection{Predictive Performance on Real-World Datasets}
\label{sec:realdata_pred}

We compare the HTNL variants with class-weighted Logistic Regression on identical train/test splits (Table~\ref{tab:realdata_11dataset} and Figure~\ref{fig:realdata_auc}). All non-collapsing HTNL variants improve over the baseline after Bonferroni correction~\citep{Bonferroni1936}. VQF performs best, increasing mean AUC by $0.074$ and outperforming Logistic Regression on all 11 datasets. ADMM-Bisect and WGL($\lambda=1$) also improve predictive performance, but their broader active sets provide no additional benefit over the sparser VQF solution. This suggests that broader feature selection does not necessarily translate into better generalization. Complementary synthetic diagnostics further show that ADMM is more sensitive than VQF to the post-hoc NCF selection threshold, with their relative F1 ranking changing across thresholds (Appendix~\ref{app:threshold_sensitivity}).

WGL further reveals the importance of optimization scale. Its default continuation schedule collapses to an empty model with AUC $=0.500$ on all 10 highly imbalanced civil-unrest datasets, whereas the scale-corrected WGL($\lambda=1$) recovers non-degenerate solutions and predictive gains. Appendix~\ref{app:wgl_sensitivity} provides the full regularization-scale analysis, while Appendix~\ref{sec:imbalance_effect} further examines this behavior across different positive-class fractions.

\begin{table}[!ht]
\centering
\small
\setlength{\tabcolsep}{3pt}
\caption{%
Predictive AUC on 11 real-world datasets using identical train/test splits. $n$ and pos\% denote the number of training instances and positive-class fraction, respectively. $\Delta$AUC denotes the improvement of VQF over Logistic Regression. The final row reports mean $\pm$ standard deviation across datasets.
}

\label{tab:realdata_11dataset}
\begin{tabular}{lrrccccc}
\toprule
\textbf{Dataset} & $n$ & pos\% & \textbf{LogReg} &
\textbf{MM/VQF} & \textbf{ADMM-Bisect} & \textbf{WGL($\lambda=1$)} &
$\Delta$AUC\\
\midrule
Flu$\star$ & 5,096 & 7.40\% & $0.649$ & $0.695$ & $0.600$ & $0.606$ & $+0.046$ \\
Ecuador & 5,916 & 0.74\% & $0.529$ & $0.604$ & $0.563$ & $0.581$ & $+0.075$ \\
El Salvador & 8,700 & 1.26\% & $0.477$ & $0.572$ & $0.559$ & $0.555$ & $+0.095$ \\
Uruguay & 10,788 & 0.82\% & $0.481$ & $0.601$ & $0.565$ & $0.582$ & $+0.120$ \\
Venezuela & 11,832 & 0.70\% & $0.463$ & $0.524$ & $0.514$ & $0.523$ & $+0.061$ \\
Chile & 12,876 & 0.49\% & $0.459$ & $0.527$ & $0.501$ & $0.503$ & $+0.068$ \\
Paraguay & 14,964 & 1.52\% & $0.518$ & $0.564$ & $0.549$ & $0.557$ & $+0.047$ \\
Colombia & 19,140 & 0.51\% & $0.523$ & $0.566$ & $0.529$ & $0.547$ & $+0.043$ \\
Argentina & 19,836 & 0.51\% & $0.453$ & $0.604$ & $0.494$ & $0.517$ & $+0.152$ \\
Brazil & 22,620 & 0.48\% & $0.531$ & $0.598$ & $0.584$ & $0.582$ & $+0.067$ \\
Mexico & 41,760 & 1.29\% & $0.590$ & $0.629$ & $0.595$ & $0.602$ & $+0.039$ \\
\midrule
\textit{Grand mean} & --- & --- & $0.516\pm0.060$ & $0.589\pm0.048$ & $0.550\pm0.037$ & $0.559\pm0.035$ & $+0.074$ \\
\bottomrule
\end{tabular}

\end{table}

Figure~\ref{fig:main_bar} summarizes the synthetic results.MM and VQF remain identical, while replacing FISTA with bisection improves ADMM F1 ($\Delta\mathrm{F1}=+0.088$, $p_{\mathrm{adj}}=2.60\times10^{-3}$). In contrast, WGL-Adaptive significantly degrades performance on both datasets. Per-dataset comparisons and runtime results are provided in Appendix~\ref{sec:perdataset}; MM/VQF achieves the highest AUC among the evaluated HTNL methods on all 10 LatAm datasets.

\subsection{KKT Screening and Computational Efficiency}
\label{app:kkt_speedup}

\begin{figure}[!ht]

\centering
\begin{subfigure}[t]{0.495\linewidth}
\centering
\includegraphics[width=\linewidth]{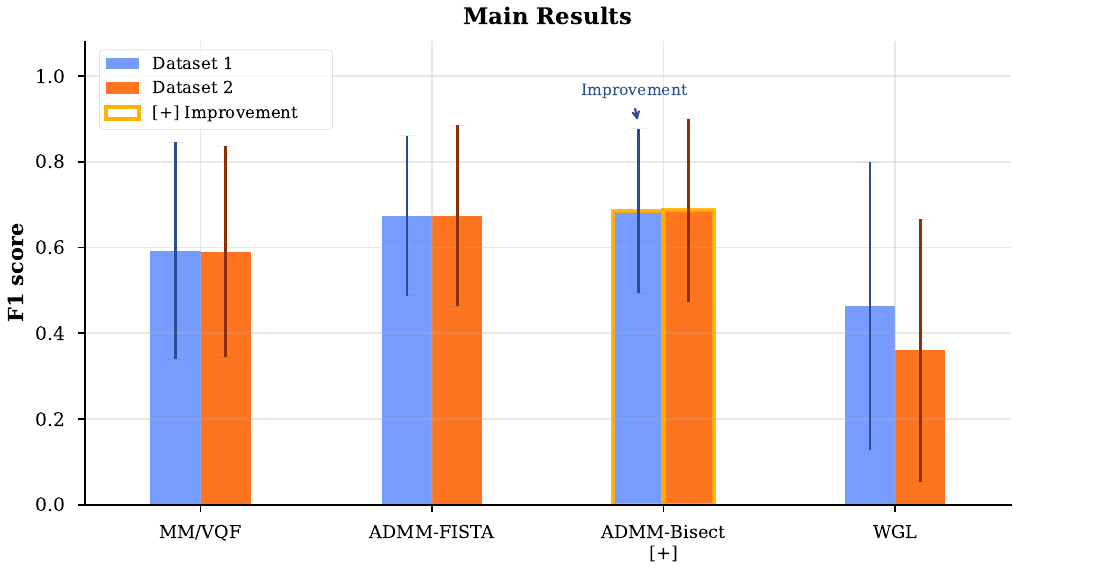}

\caption{Synthetic F1.}
\label{fig:main_bar}
\end{subfigure}\hspace{0.005\linewidth}%
\begin{subfigure}[t]{0.495\linewidth}
\centering
\includegraphics[width=\linewidth]{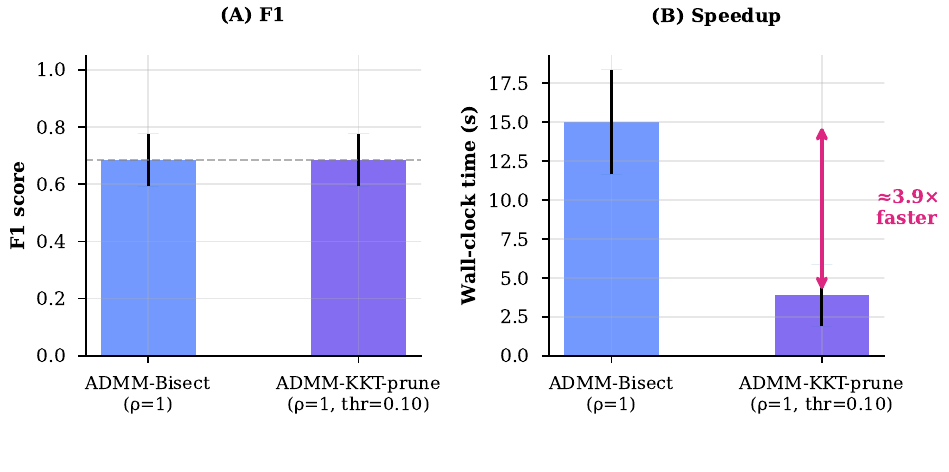}

\caption{KKT pruning: F1 and runtime.}
\label{fig:kkt_speedup}
\end{subfigure}

\caption{Controlled optimization diagnostics. \textbf{Left:} Bisection improves ADMM F1, while MM and VQF are identical. \textbf{Right:} KKT pruning preserves F1 while reducing runtime.}
\label{fig:controlled_main}

\end{figure}

We further evaluate the computational effect of combining KKT pruning with the bisection proximal solver. Figure~\ref{fig:kkt_speedup} compares the baseline ADMM configuration at $\rho=1$ with ADMM using KKT pruning and bisection at the same $\rho$, with pruning threshold $0.10$. The two configurations achieve nearly identical synthetic F1, while the KKT-pruned bisection variant substantially reduces wall-clock time. In this diagnostic, runtime decreases from approximately $15$\,s to $4$\,s, corresponding to an approximately $3.9\times$ speedup. This result shows that the combined pruning and bisection implementation can reduce computational cost without an observable loss in F1 under this experimental condition.

\subsection{Sensitivity to the Active-Set Admission Threshold}
\label{app:threshold_sensitivity}

\begin{figure}[!ht]

\centering
\includegraphics[width=0.9\linewidth]{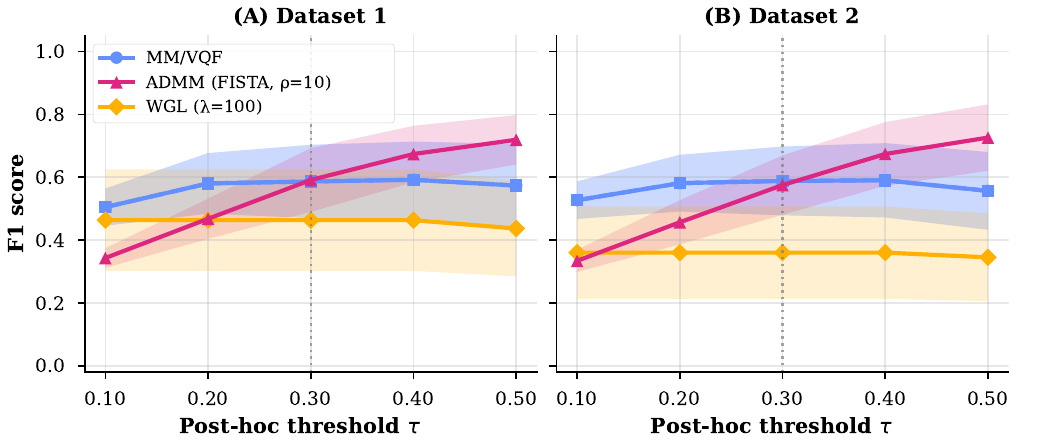}

\caption{Synthetic F1 across active-set admission thresholds $\tau_{\mathrm{KKT}}$. Shaded bands show uncertainty across paired random seeds.}
\label{fig:f1_vs_rel}

\end{figure}

Figure~\ref{fig:f1_vs_rel} examines sensitivity to the active-set admission threshold $\tau_{\mathrm{KKT}}$ in Equation~\ref{eq:active_set_update}, which determines whether an inactive NCF is admitted to the active set based on its KKT score. ADMM is more sensitive to this threshold than VQF: changing $\tau_{\mathrm{KKT}}$ alters the admitted candidates and can reverse their relative F1 ranking, whereas VQF remains comparatively stable. This result shows that optimization comparisons can depend on the active-set admission threshold and motivates evaluation across multiple threshold settings.

\section{Conclusion}

\label{sec:conclusion}

We studied four HTNL optimization methods for Numerical Conjunctive Feature selection. Our analysis shows that MM and VQF are equivalent under matched updates, ADMM-Bisect corrects the inexact proximal step, and all methods remain limited by the shared active-set KKT rule. Across 11 real-world datasets, non-collapsing HTNL variants improve over Logistic Regression, while WGL requires careful regularization scaling. Overall, improving the inner solver helps optimization, but overcoming the remaining active-set limit requires a better candidate-expansion strategy.

\subsection*{AI Use Statement}
AI tools assisted with LaTeX formatting, organization of the main text and appendices, and layout checks. The original theoretical statements and reported experimental were retained.
\label{main:end}

\bibliographystyle{iclr2027_conference}
\bibliography{references}
\clearpage
% Bibliography generated with the official ICLR 2027 .bst, then embedded.

\appendix

\section{Related Work}

\label{sec:related}

\paragraph{Spatial event forecasting and conjunctive precursor learning.}
Forecasting societal events from social media has been studied in civil unrest, disease outbreaks, crime prediction, and other spatial-temporal settings~\citep{Ramakrishnan2014,Gerber2014,Compton2014,Lamb2013,Achrekar2011}. Many systems use keyword counts, topic volumes, burstiness, or spatial density signals as predictive features. These features are scalable, but they mostly model single-factor evidence. NCFs instead represent interactions among primitive signals, making them closer to rule-learning and precursor-discovery methods~\citep{Friedman2008,Ruckert2006,Cheng2007,Cong2005}. HTNL~\citep{Zhao2019htnl} extends this direction by learning numerical conjunctions directly from an exponential lattice while also modeling geographic structure.

\paragraph{Structured sparsity and hierarchical selection.}
HTNL belongs to a broad family of sparsity-inducing methods, including lasso~\citep{Tibshirani1996}, group lasso~\citep{Yuan2006,Meier2008}, overlapping group lasso~\citep{Jacob2009}, sparse-group lasso~\citep{Simon2013}, hierarchical sparsity~\citep{Bach2009,Jenatton2011,Bach2012}, and multi-task feature selection~\citep{Argyriou2008,Obozinski2010,Kim2010}. These methods provide the modeling language for group and hierarchy selection, but HTNL adds two complications that are central to this paper: an inner geographic group norm nested inside an outer NCF hierarchy, and an active-set search over exponentially many candidate conjunctions. Multiple kernel learning is also related through its use of norm-controlled component selection~\citep{Gonen2011,Kloft2011}, but it does not address active-set NCF recovery.

\paragraph{Proximal optimization, ADMM, and active sets.}
The HTNL objective combines a smooth convex loss with a nonsmooth hierarchical regularizer, making it closely related to proximal algorithms, splitting methods, and working-set optimization~\citep{Beck2009,Boyd2011,EcksteinBertsekas1992,Parikh2014,Bach2012}. ADMM is a standard tool for such composite objectives, but support recovery can be sensitive to inexact inner proximal solves when sparsity is the target. Active-set and path methods for lasso-style problems reduce computation by adding variables that violate optimality conditions~\citep{Osborne2000,Tibshirani2012,Friedman2010}. Our contribution is to diagnose how this active-set mechanism behaves in the NCF lattice: improving the inner solver can remove numerical artifacts, but the KKT admission rule itself can still limit which high-order conjunctions are reachable.

\paragraph{Neural and language-model event forecasting.}
Recent event forecasting work increasingly uses deep spatial-temporal representations, graph neural networks, and language models~\citep{Deng2021survey,Deng2024kdd,Chowdhury2024detective,Liao2024gentkg,Li2024mmforecast}. These methods are powerful predictors, but they usually return latent representations, generated events, or end-task predictions rather than sparse, inspectable precursor rules. HTNL is complementary: it targets transparent NCF discovery, and our paper focuses on making the structured-sparsity optimization reliable enough to support that interpretability goal.

\section{Detailed Optimization Method Formulations}\label{app:optimizer_details}

\subsection{Direct ADMM optimization on a fixed active set}\label{sec:direct_admm}
The variational quadratic form in Section~\ref{sec:vqf_method} converts the HTNL regularizer into a weighted ridge problem after auxiliary variables are fixed. We next consider a direct ADMM solver that keeps the original nonsmooth two-level group lasso structure. The key idea is to split each active NCF block into auxiliary consensus variables, so that the smooth loss is optimized in the $W$-update and the nested group penalty is handled in the $Z$-update.

Let $U\subseteq\mathcal V$ be the current active set of numerical conjunctive features (NCFs). We write the restricted parameter collection as
\begin{equation} W_U=\{W_{j,u}:u\in U,\ j\in S\}, \label{eq:admm-active-parameters} \end{equation}
where $u$ indexes an NCF, $j$ indexes a spatial location, and $S$ denotes the full set of spatial locations. Let $\mathcal G$ be the set of geographical groups, where each $g\in\mathcal G$ is a subset of $S$. For each $u\in U$ and $g\in\mathcal G$, define the block extraction operator
\begin{equation} P_{u,g}W_U=\{W_{j,u}:j\in g\}. \label{eq:block-extraction} \end{equation}
Thus, $P_{u,g}W_U$ extracts the parameter block associated with NCF $u$ and geographical group $g$.

On the active set $U$, define the restricted descendant set
\begin{equation} D_U(v)=D(v)\cap U. \label{eq:restricted-descendant} \end{equation}
The direct nonsmooth optimization problem is
\begin{equation} \min_{W_U}\mathcal L(W_U)+\lambda\sum_{v\in U}d_v\left(\sum_{u\in D_U(v)}\left(\sum_{g\in\mathcal G}\|P_{u,g}W_U\|_2\right)^p\right)^{1/p}. \label{eq:restricted-nonsmooth-problem} \end{equation}
The difficulty in Equation~\eqref{eq:restricted-nonsmooth-problem} is that the regularizer contains both the geographical group norm $\sum_{g\in\mathcal G}\|P_{u,g}W_U\|_2$ and the hierarchical coupling over $u\in D_U(v)$. Therefore, a direct proximal step over $W_U$ is not generally available in closed form.

To apply ADMM, we introduce one auxiliary copy for every block appearing in every hierarchical regularization term:
\begin{equation} Z_{v,u,g}=P_{u,g}W_U,\qquad v\in U,\ u\in D_U(v),\ g\in\mathcal G. \label{eq:admm-copy-definition} \end{equation}
The constrained reformulation of Equation~\eqref{eq:restricted-nonsmooth-problem} is
\begin{equation} \begin{aligned} &\min_{W_U,Z}\mathcal L(W_U)+\lambda\sum_{v\in U}d_v\left(\sum_{u\in D_U(v)}\left(\sum_{g\in\mathcal G}\|Z_{v,u,g}\|_2\right)^p\right)^{1/p}\\ \mathrm{s.t.} \quad &Z_{v,u,g}=P_{u,g}W_U,\ v\in U,\ u\in D_U(v),\ g\in\mathcal G. \label{eq:admm-constrained-problem} \end{aligned} \end{equation}
This splitting places the smooth loss on $W_U$ and the nonsmooth double group regularizer on $Z$, with the two parts connected only through linear consensus constraints.

Using scaled dual variables $Y_{v,u,g}$ and penalty parameter $\rho>0$, the scaled augmented Lagrangian is
\begin{equation} \begin{aligned} \mathcal L_\rho(W_U,Z,Y)=\mathcal L(W_U)&+\lambda\sum_{v\in U}d_v\left(\sum_{u\in D_U(v)}\left(\sum_{g\in\mathcal G}\|Z_{v,u,g}\|_2\right)^p\right)^{1/p}\\&+\frac{\rho}{2}\sum_{v\in U}\sum_{u\in D_U(v)}\sum_{g\in\mathcal G}\|P_{u,g}W_U-Z_{v,u,g}+Y_{v,u,g}\|_2^2. \label{eq:admm-augmented-lagrangian} \end{aligned} \end{equation}

ADMM then alternates the following three updates. First, the $W_U$-update solves
\begin{equation} W_U^{k+1}=\arg\min_{W_U}\mathcal L(W_U)+\frac{\rho}{2}\sum_{v\in U}\sum_{u\in D_U(v)}\sum_{g\in\mathcal G}\|P_{u,g}W_U-Z_{v,u,g}^k+Y_{v,u,g}^k\|_2^2. \label{eq:admm-w-update-rigid} \end{equation}
This subproblem contains no nonsmooth group norm. If $\mathcal L$ is squared loss, Equation~\eqref{eq:admm-w-update-rigid} is a quadratic least-squares problem. If $\mathcal L$ is logistic or another smooth convex loss, Equation~\eqref{eq:admm-w-update-rigid} can be solved by Newton, L-BFGS, or an inner gradient method. The quadratic term in Equation~\eqref{eq:admm-w-update-rigid} is an ADMM consensus penalty, not a variational ridge reformulation of the original regularizer.

Second, define
\begin{equation} Q_{v,u,g}^{k+1}=P_{u,g}W_U^{k+1}+Y_{v,u,g}^k. \label{eq:admm-q-rigid} \end{equation}
The $Z$-update is
\begin{equation} \begin{aligned} Z^{k+1}&=\arg\min_Z\lambda\sum_{v\in U}d_v\left(\sum_{u\in D_U(v)}\left(\sum_{g\in\mathcal G}\|Z_{v,u,g}\|_2\right)^p\right)^{1/p}\\&+\frac{\rho}{2}\sum_{v\in U}\sum_{u\in D_U(v)}\sum_{g\in\mathcal G}\|Z_{v,u,g}-Q_{v,u,g}^{k+1}\|_2^2. \label{eq:admm-z-update-rigid} \end{aligned} \end{equation}
Because the regularizer is separable over the ancestor index $v$, Equation~\eqref{eq:admm-z-update-rigid} decomposes into independent subproblems:
\begin{equation} Z_v^{k+1}=\arg\min_{Z_v}\lambda d_v\left(\sum_{u\in D_U(v)}\left(\sum_{g\in\mathcal G}\|Z_{v,u,g}\|_2\right)^p\right)^{1/p}+\frac{\rho}{2}\sum_{u\in D_U(v)}\sum_{g\in\mathcal G}\|Z_{v,u,g}-Q_{v,u,g}^{k+1}\|_2^2. \label{eq:admm-z-v-subproblem} \end{equation}
Equivalently, define
\begin{equation} \Omega_v(Z_v)=\left(\sum_{u\in D_U(v)}\left(\sum_{g\in\mathcal G}\|Z_{v,u,g}\|_2\right)^p\right)^{1/p}. \label{eq:local-double-group-regularizer} \end{equation}
Then Equation~\eqref{eq:admm-z-v-subproblem} can be written as the proximal update
\begin{equation} Z_v^{k+1}=\operatorname{prox}_{(\lambda d_v/\rho)\Omega_v}(Q_v^{k+1}). \label{eq:admm-z-prox-rigid} \end{equation}
This proximal operator is not the ordinary group soft-thresholding operator, because $\Omega_v$ contains two coupled levels: an inner sum of geographical $\ell_2$ group norms over $g\in\mathcal G$, followed by an outer $\ell_p$ aggregation over $u\in D_U(v)$. Thus, unless the special choice of $p$ and group structure gives a closed form, Equation~\eqref{eq:admm-z-prox-rigid} must itself be solved by an inner proximal, dual-projection, or primal-dual splitting routine.

Third, the scaled dual variables are updated by
\begin{equation} Y_{v,u,g}^{k+1}=Y_{v,u,g}^k+P_{u,g}W_U^{k+1}-Z_{v,u,g}^{k+1}. \label{eq:admm-y-update-rigid} \end{equation}
This update accumulates the violation of the consensus constraint in Equation~\eqref{eq:admm-copy-definition}. The primal residual for the consensus constraint is
\begin{equation} r_{v,u,g}^{k+1}=P_{u,g}W_U^{k+1}-Z_{v,u,g}^{k+1}. \label{eq:admm-primal-residual} \end{equation}
The corresponding dual residual is
\begin{equation} s^{k+1}=\rho\sum_{v\in U}\sum_{u\in D_U(v)}\sum_{g\in\mathcal G}P_{u,g}^{\top}\left(Z_{v,u,g}^{k+1}-Z_{v,u,g}^k\right), \label{eq:admm-dual-residual} \end{equation}
where $P_{u,g}^{\top}$ is the adjoint block insertion operator, which places a block back into the coordinates of $W_U$. The ADMM inner loop is terminated when
\begin{equation} \|r^{k+1}\|_2\le\epsilon_{\mathrm{pri}},\qquad \|s^{k+1}\|_2\le\epsilon_{\mathrm{dual}}. \label{eq:admm-stopping} \end{equation}
ADMM directly optimizes the original nonsmooth double group lasso on a fixed active set $U$ by splitting each block $P_{u,g}W_U$ into copies $Z_{v,u,g}$. The $W_U$-update in Equation~\eqref{eq:admm-w-update-rigid} solves a smooth loss plus a quadratic consensus penalty, while the $Z$-update in Equation~\eqref{eq:admm-z-update-rigid} keeps the original double group lasso structure and computes its proximal map. For the full HTNL problem, this ADMM routine should still be embedded inside an outer active-set procedure: after solving Equation~\eqref{eq:restricted-nonsmooth-problem} on $U$, one checks the optimality violation over inactive NCFs in the lattice, adds violating descendants to $U$, and repeats until no inactive NCF violates the optimality condition.

\subsection{Reduction from two-level group lasso to weighted one-level group lasso}\label{sec:wgl_reduction}

The ADMM formulation in Section~\ref{sec:direct_admm} preserves the original two-level nonsmooth structure but requires solving a nested proximal problem. We now present an alternative reduction that converts the local $\ell_p$ hierarchy into a weighted one-level group lasso problem. The main idea is to use the dual representation of the outer $\ell_p$ norm: after fixing the dual weights, the two-level penalty becomes an ordinary weighted group-lasso penalty over geographic blocks. The hierarchy is restored by updating the dual weights in closed form.

We use the same fixed active set $U$, parameter collection $W_U$, block extraction operator $P_{u,g}$, and restricted descendant set $D_U(v)$ defined in~\eqref{eq:admm-active-parameters}, \eqref{eq:block-extraction}, \eqref{eq:restricted-descendant}. Under this notation, the restricted two-level group lasso problem is the same as Equation~\eqref{eq:restricted-nonsmooth-problem}.
Define the inner geographical group magnitude
\begin{equation} r_{v,u}(W_U)=\sum_{g\in\mathcal G}\|P_{u,g}W_U\|_2,\qquad v\in U,\ u\in D_U(v). \label{eq:rvu-definition} \end{equation}
Then the local hierarchical penalty associated with ancestor $v$ is
\begin{equation} \Omega_v(W_U)=\left(\sum_{u\in D_U(v)}r_{v,u}(W_U)^p\right)^{1/p}. \label{eq:omegav-definition} \end{equation}
Let $q$ be the dual exponent of $p$:
\begin{equation} \frac{1}{p}+\frac{1}{q}=1. \label{eq:dual-exponent} \end{equation}
By the dual representation of the $\ell_p$ norm, Equation~\eqref{eq:omegav-definition} can be written as
\begin{equation} \Omega_v(W_U)=\max_{\alpha_v\in\mathcal A_v}\sum_{u\in D_U(v)}\alpha_{v,u}r_{v,u}(W_U), \label{eq:lp-dual-representation} \end{equation}
where
\begin{equation} \mathcal A_v=\left\{\alpha_v\in\mathbb R_+^{|D_U(v)|}:\|\alpha_v\|_q\le 1\right\}. \label{eq:alpha-feasible-set} \end{equation}
Substituting Equation~\eqref{eq:lp-dual-representation} into Equation~\eqref{eq:restricted-nonsmooth-problem} gives the equivalent saddle-point formulation
\begin{equation} \min_{W_U}\max_{\{\alpha_v\in\mathcal A_v\}_{v\in U}}\mathcal L(W_U)+\lambda\sum_{v\in U}d_v\sum_{u\in D_U(v)}\alpha_{v,u}\sum_{g\in\mathcal G}\|P_{u,g}W_U\|_2. \label{eq:saddle-point-one-layer} \end{equation}
For fixed $\alpha=\{\alpha_v\}_{v\in U}$, collect the coefficients that multiply the same block norm $\|P_{u,g}W_U\|_2$. Define the active ancestor set
\begin{equation} A_U(u)=\{v\in U:u\in D_U(v)\}. \label{eq:active-ancestor-set} \end{equation}
The collapsed block weight is
\begin{equation} c_{u,g}(\alpha)=\lambda\sum_{v\in A_U(u)}d_v\alpha_{v,u}. \label{eq:collapsed-block-weight} \end{equation}
Since $c_{u,g}(\alpha)$ does not depend on $g$ under the penalty in Equation~\eqref{eq:restricted-nonsmooth-problem}, one may also write $c_u(\alpha)=\lambda\sum_{v\in A_U(u)}d_v\alpha_{v,u}$. We keep the notation $c_{u,g}(\alpha)$ to match the blockwise group-lasso form. For fixed $\alpha$, Equation~\eqref{eq:saddle-point-one-layer} reduces to the weighted one-level group lasso problem
\begin{equation} \min_{W_U}\mathcal L(W_U)+\sum_{u\in U}\sum_{g\in\mathcal G}c_{u,g}(\alpha)\|P_{u,g}W_U\|_2. \label{eq:weighted-one-level-group-lasso} \end{equation}

\textbf{One-level group-lasso update.}
Assume that $\mathcal L$ is differentiable and has $L_{\mathcal L}$-Lipschitz gradient. Given a stepsize $\eta\le 1/L_{\mathcal L}$, the proximal-gradient update for Equation~\eqref{eq:weighted-one-level-group-lasso} is
\begin{equation} \widetilde W_U^k=W_U^k-\eta\nabla\mathcal L(W_U^k). \label{eq:gradient-step-one-level} \end{equation}
Then
\begin{equation} W_U^{k+1}=\operatorname{prox}_{\eta\sum_{u\in U}\sum_{g\in\mathcal G}c_{u,g}(\alpha)\|P_{u,g}\cdot\|_2}(\widetilde W_U^k). \label{eq:prox-step-one-level} \end{equation}
If the groups $\{P_{u,g}\}_{u,g}$ are non-overlapping, Equation~\eqref{eq:prox-step-one-level} has the block soft-thresholding solution
\begin{equation} P_{u,g}W_U^{k+1}=\left(1-\frac{\eta c_{u,g}(\alpha)}{\|P_{u,g}\widetilde W_U^k\|_2}\right)_+P_{u,g}\widetilde W_U^k. \label{eq:block-soft-thresholding} \end{equation}
If the geographical groups overlap, Equation~\eqref{eq:weighted-one-level-group-lasso} remains a one-level overlapping group lasso problem and can be solved by a standard overlapping-group solver, such as latent group lasso, ADMM, or a dual projected method.

\textbf{Outer dual-weight update.}
Given $W_U$, the maximizer in Equation~\eqref{eq:lp-dual-representation} has a closed form when $1<p<\infty$. For each $v\in U$ and $u\in D_U(v)$,
\begin{equation} \alpha_{v,u}^{\star}(W_U)=\frac{r_{v,u}(W_U)^{p-1}}{\left(\sum_{u'\in D_U(v)}r_{v,u'}(W_U)^p\right)^{(p-1)/p}}. \label{eq:alpha-closed-form} \end{equation}
If $r_{v,u}(W_U)=0$ for all $u\in D_U(v)$, any $\alpha_v\in\mathcal A_v$ is optimal. In implementation, one can replace $r_{v,u}(W_U)$ by $r_{v,u}(W_U)+\epsilon$ for a small $\epsilon>0$.

\begin{algorithm}[t]
\caption{Two-Level Group Lasso via Weighted One-Level Group Lasso}
\label{alg:two-level-to-one-level}
\begin{algorithmic}[1]
\REQUIRE Active set $U$, loss $\mathcal L$, groups $\mathcal G$, descendant sets $\{D_U(v)\}_{v\in U}$, weights $\{d_v\}_{v\in U}$, regularization parameter $\lambda$, exponent $p$, tolerance $\varepsilon$.
\ENSURE Optimized parameter $W_U$.
\STATE Initialize $W_U^0$ and $\alpha^0=\{\alpha_v^0\}_{v\in U}$ with $\alpha_v^0\in\mathcal A_v$.
\FOR{$t=0,1,2,\ldots$}
\STATE Compute collapsed one-level group weights
\begin{equation} c_{u,g}^t=\lambda\sum_{v\in A_U(u)}d_v\alpha_{v,u}^t,\qquad u\in U,\ g\in\mathcal G. \label{eq:alg-collapsed-weight} \end{equation}
\STATE Solve the weighted one-level group lasso subproblem
\begin{equation} W_U^{t+1} = \arg\min_{W_U}\mathcal L(W_U)+\sum_{u\in U}\sum_{g\in\mathcal G}c_{u,g}^t\|P_{u,g}W_U\|_2. \label{eq:alg-one-level-subproblem} \end{equation}
\STATE Compute inner group magnitudes
\begin{equation} r_{v,u}^{t+1}=\sum_{g\in\mathcal G}\|P_{u,g}W_U^{t+1}\|_2,\qquad v\in U,\ u\in D_U(v). \label{eq:alg-r-update} \end{equation}
\STATE Update outer dual weights
\begin{equation} \alpha_{v,u}^{t+1}=\frac{(r_{v,u}^{t+1}+\epsilon)^{p-1}}{\left(\sum_{u'\in D_U(v)}(r_{v,u'}^{t+1}+\epsilon)^p\right)^{(p-1)/p}},\qquad v\in U,\ u\in D_U(v). \label{eq:alg-alpha-update} \end{equation}
\IF{$\|W_U^{t+1}-W_U^t\|_F\le\varepsilon$}
\STATE \textbf{break}
\ENDIF
\ENDFOR
\RETURN $W_U^{t+1}$.
\end{algorithmic}
\end{algorithm}

The transformation in Equation~\eqref{eq:saddle-point-one-layer} is exact only when the outer dual variables $\alpha$ are optimized together with $W_U$. For fixed $\alpha$, the $W_U$-subproblem in Equation~\eqref{eq:weighted-one-level-group-lasso} is a weighted one-level group lasso problem. The update in Equation~\eqref{eq:alpha-closed-form} restores the original outer $\ell_p$ hierarchy by choosing the dual weights that exactly represent the local $\ell_p$ norm in Equation~\eqref{eq:omegav-definition}. Therefore, the full procedure alternates between a standard one-level group-lasso solve and a closed-form update of the outer hierarchy weights.

The reduction above is performed on a fixed active set $U$. It does not remove the exponential number of candidate NCFs in the full feature conjunction lattice. Therefore, the method should be embedded inside an outer active-set procedure: solve the restricted problem on $U$ using Algorithm~\ref{alg:two-level-to-one-level}, check the optimality violation over inactive descendants in the lattice, add violating NCFs to $U$, and repeat until no inactive NCF violates the optimality condition.

\section{Proofs}\label{app:proofs}

\subsection{Proof of Lemma~\ref{lemma:1}}\label{app:lemma1}

\begin{proof}
We use the following variational identity following from Hölder's inequality~\citep{hardy1952inequalities}: for any $a_1,\ldots,a_m\ge 0$ and $r>0$,
\begin{equation} \left(\sum_{i=1}^m a_i^{r/(r+1)}\right)^{(r+1)/r}=\min_{\theta\in\Theta_{m,r}}\sum_{i=1}^m\frac{a_i}{\theta_i}, \label{eq:variational-identity} \end{equation}
with equality attained when $\theta_i$ is proportional to $a_i^{1/(r+1)}$.

First, define the squared group magnitude $A_{u,g}(W)=\sum_{j\in g}\|W_{j,u}\|_2^2$. Then $\left\|\{W_{j,u}\}_{j\in g}\right\|_2=A_{u,g}(W)^{1/2}$. Applying~\eqref{eq:variational-identity} with $r=1$ to the geographical group term gives
\begin{equation} \left(\sum_{g=1}^G\left\|\{W_{j,u}\}_{j\in g}\right\|_2\right)^2=\min_{\mu_{u,v}\in\Theta_{G,1}}\sum_{g=1}^G\frac{1}{\mu_{u,v,g}}\sum_{j\in g}\|W_{j,u}\|_2^2. \label{eq:geo-group-quadratic} \end{equation}

Next, define
\begin{equation} B_u(W)=\sum_{g=1}^G\left\|\{W_{j,u}\}_{j\in g}\right\|_2. \label{eq:bu-definition} \end{equation}
For each fixed $v$, the squared feature-hierarchy term in~\eqref{eq:omega-two-level} is
\begin{equation} \left(\sum_{u\in D(v)}B_u(W)^p\right)^{2/p}. \label{eq:feature-hierarchy-squared} \end{equation}
Since $1<p\le 2$, set $\bar p=p/(2-p)$. Applying~\eqref{eq:variational-identity} to~\eqref{eq:feature-hierarchy-squared} yields
\begin{equation} \left(\sum_{u\in D(v)}B_u(W)^p\right)^{2/p}=\min_{\lambda_v\in\Theta_{|D(v)|,\bar p}}\sum_{u\in D(v)}\frac{B_u(W)^2}{\lambda_{v,u}}. \label{eq:feature-hierarchy-variational} \end{equation}
Substituting~\eqref{eq:geo-group-quadratic} into~\eqref{eq:feature-hierarchy-variational} gives
\begin{equation} \left(\sum_{u\in D(v)}B_u(W)^p\right)^{2/p}=\min_{\lambda_v,\mu}\sum_{u\in D(v)}\frac{1}{\lambda_{v,u}}\sum_{g=1}^G\frac{1}{\mu_{u,v,g}}\sum_{j\in g}\|W_{j,u}\|_2^2. \label{eq:feature-geo-quadratic} \end{equation}
Now define
\begin{equation} C_v(W)=d_v\left(\sum_{u\in D(v)}B_u(W)^p\right)^{1/p}. \label{eq:cv-definition} \end{equation}
Then $\Omega(W)=\sum_{v\in\mathcal V}C_v(W)$. Applying~\eqref{eq:variational-identity} with $r=1$ gives
\begin{equation} \Omega(W)^2=\left(\sum_{v\in\mathcal V}C_v(W)\right)^2=\min_{\gamma\in\Theta_{|\mathcal V|,1}}\sum_{v\in\mathcal V}\frac{C_v(W)^2}{\gamma_v}. \label{eq:outer-variational} \end{equation}
By~\eqref{eq:cv-definition},
\begin{equation} C_v(W)^2=d_v^2\left(\sum_{u\in D(v)}B_u(W)^p\right)^{2/p}. \label{eq:cv-squared} \end{equation}
Substituting~\eqref{eq:feature-geo-quadratic} and~\eqref{eq:cv-squared} into~\eqref{eq:outer-variational} yields
\begin{equation} \Omega(W)^2=\min_{\gamma,\lambda,\mu}\sum_{v\in\mathcal V}\frac{d_v^2}{\gamma_v}\sum_{u\in D(v)}\frac{1}{\lambda_{v,u}}\sum_{g=1}^G\frac{1}{\mu_{u,v,g}}\sum_{j\in g}\|W_{j,u}\|_2^2, \end{equation}
which is exactly~\eqref{eq:omega-quadratic-representation}. For fixed $\gamma,\lambda,\mu$, collect all terms in~\eqref{eq:omega-quadratic-representation} that multiply the same block $\sum_{j\in g}\|W_{j,u}\|_2^2$. Define
\begin{equation} \Psi_{u,g}(\gamma,\lambda,\mu)=\left(\sum_{v\in A(u)}\frac{d_v^2}{\gamma_v\lambda_{v,u}\mu_{u,v,g}}\right)^{-1}. \label{eq:psi-definition} \end{equation}
Then~\eqref{eq:omega-quadratic-representation} can be rewritten as
\begin{equation} \Omega(W)^2=\min_{\gamma,\lambda,\mu}\sum_{u\in\mathcal V}\sum_{g=1}^G\Psi_{u,g}^{-1}(\gamma,\lambda,\mu)\sum_{j\in g}\|W_{j,u}\|_2^2. \label{eq:weighted-ridge-form} \end{equation}
Therefore, conditional on $\gamma,\lambda,\mu$, the two-level group lasso regularizer in~\eqref{eq:omega-two-level} reduces to a weighted ridge penalty over $W$.
\end{proof}

\subsection{\texorpdfstring{Proof of Proposition~\ref{prop:equiv} (VQF $\equiv$ MM)}{Proof of Proposition: VQF equivalent to MM}}
\label{app:prop1}

We prove that the MM solver and the VQF solver generate identical $W$-iterates when they use the same exact variational weights. The proof follows from the fact that both methods reduce the HTNL regularizer to the same weighted-ridge subproblem after the auxiliary variables are fixed.

% \begin{theorem}[Equivalence of MM and VQF]
% \label{thm:vqf_original_equiv}
% Let $\Omega(W)$ be the two-level HTNL regularizer in~\eqref{eq:omega-two-level}, and let its variational quadratic representation be given by~\eqref{eq:omega-quadratic-representation}. Suppose MM and VQF start from the same initialization $W^{(0)}$ and update the auxiliary variables $(\gamma,\lambda,\mu)$ by the same exact minimization of the variational representation at the current iterate. Then, for every iteration $t\ge 0$, the two methods solve the same weighted-ridge subproblem and produce identical iterates:
% \begin{equation}
% \end{theorem}

\begin{proof}
Define the variational quadratic form
\begin{equation} R(W;\gamma,\lambda,\mu)=\sum_{v\in\mathcal{V}}\frac{d_v^2}{\gamma_v}\sum_{u\in D(v)}\frac{1}{\lambda_{v,u}}\sum_{g=1}^{G}\frac{1}{\mu_{u,v,g}}\sum_{j\in g}\|W_{j,u}\|_2^2. \label{eq:rqf_definition_app} \end{equation}
By the variational identity in~\eqref{eq:omega-quadratic-representation}, the squared HTNL regularizer satisfies
\begin{equation} \Omega(W)^2=\min_{\gamma,\lambda,\mu}R(W;\gamma,\lambda,\mu), \label{eq:omega_min_rqf_app} \end{equation}
where $\gamma\in\Theta_{|\mathcal{V}|,1}$, $\lambda_v\in\Theta_{|D(v)|,\bar p}$, $\mu_{u,v}\in\Theta_{G,1}$, and $\bar p=p/(2-p)$.

At iteration $t$, both MM and VQF compute the same auxiliary variables by minimizing the variational form at the current iterate:
\begin{equation} (\gamma^{(t+1)},\lambda^{(t+1)},\mu^{(t+1)})\in\arg\min_{\gamma,\lambda,\mu}R(W^{(t)};\gamma,\lambda,\mu). \label{eq:aux_update_same_app} \end{equation}
For fixed $(\gamma^{(t+1)},\lambda^{(t+1)},\mu^{(t+1)})$, the regularizer becomes a quadratic penalty in $W$. Collecting all terms that multiply the same block $\sum_{j\in g}\|W_{j,u}\|_2^2$, define
\begin{equation} \Psi_{u,g}^{(t+1)}=\left(\sum_{v\in A(u)}\frac{d_v^2}{\gamma_v^{(t+1)}\lambda_{v,u}^{(t+1)}\mu_{u,v,g}^{(t+1)}}\right)^{-1}. \label{eq:psi_iter_app} \end{equation}
Then the fixed-auxiliary regularizer is
\begin{equation} R(W;\gamma^{(t+1)},\lambda^{(t+1)},\mu^{(t+1)})=\sum_{u\in\mathcal{V}}\sum_{g=1}^{G}\left(\Psi_{u,g}^{(t+1)}\right)^{-1}\sum_{j\in g}\|W_{j,u}\|_2^2. \label{eq:fixed_aux_ridge_app} \end{equation}
Thus, the $W$-update solved by MM is
\begin{equation} W_{\mathrm{MM}}^{(t+1)}=\arg\min_W\mathcal{L}(W)+\lambda_{\mathrm{reg}}\sum_{u\in\mathcal{V}}\sum_{g=1}^{G}\left(\Psi_{u,g}^{(t+1)}\right)^{-1}\sum_{j\in g}\|W_{j,u}\|_2^2. \label{eq:mm_update_correct_app} \end{equation}
The VQF solver uses the same quadratic representation in~\eqref{eq:omega_min_rqf_app}. With the same auxiliary variables from~\eqref{eq:aux_update_same_app}, its $W$-update is
\begin{equation} W_{\mathrm{VQF}}^{(t+1)}=\arg\min_W\mathcal{L}(W)+\lambda_{\mathrm{reg}}\sum_{u\in\mathcal{V}}\sum_{g=1}^{G}\left(\Psi_{u,g}^{(t+1)}\right)^{-1}\sum_{j\in g}\|W_{j,u}\|_2^2. \label{eq:vqf_update_correct_app} \end{equation}
The optimization problems in~\eqref{eq:mm_update_correct_app} and~\eqref{eq:vqf_update_correct_app} are identical. Therefore,
\begin{equation} W_{\mathrm{MM}}^{(t+1)}=W_{\mathrm{VQF}}^{(t+1)}. \label{eq:one_step_equiv_app} \end{equation}

Since both methods start from the same initialization $W^{(0)}$, the same auxiliary variables are computed at the first iteration, and the same $W$-update is solved. By induction, if $W_{\mathrm{MM}}^{(t)}=W_{\mathrm{VQF}}^{(t)}$, then~\eqref{eq:aux_update_same_app} gives the same auxiliary variables and~\eqref{eq:one_step_equiv_app} gives $W_{\mathrm{MM}}^{(t+1)}=W_{\mathrm{VQF}}^{(t+1)}$. Hence the two methods generate identical iterates for all $t\ge 0$.
\end{proof}

The proposition proves mathematical equivalence under exact auxiliary-variable updates and the same numerical convention for zero-norm blocks. In our implementation, MM and VQF call the same weighted-ridge solver after constructing the same block weights $\Psi_{u,g}^{(t+1)}$. This explains the observed bit-exact agreement of their selected NCF sets in the real-data experiments.
% =============================================================================
% Appendix B: Proof of Proposition 2 (Inexact ADMM Residual)
% =============================================================================

\subsection{Proof of Proposition~\ref{prop:admm_residual} (Inexact ADMM Residual)}
\label{app:prop2}

\begin{proof}
Let
\begin{equation} \mathcal{F}_{\rho}(Z)=\lambda\Omega^2(Z)+\frac{\rho}{2}\|Z-Q\|_2^2. \label{eq:prox_objective_proof} \end{equation}
The regularizer $\Omega^2$ is convex, and the quadratic term is $\rho$-strongly convex. Therefore, $\mathcal{F}_{\rho}$ is also $\rho$-strongly convex. By strong convexity, for any $Z$ and the unique minimizer $Z^*$,
\begin{equation} \mathcal{F}_{\rho}(Z)-\mathcal{F}_{\rho}(Z^*)\ge \frac{\rho}{2}\|Z-Z^*\|_2^2. \label{eq:strong_convexity_gap} \end{equation}
Substituting $Z=Z^{(K)}$ gives
\begin{equation} \varepsilon_K=\mathcal{F}_{\rho}(Z^{(K)})-\mathcal{F}_{\rho}(Z^*)\ge \frac{\rho}{2}\|Z^{(K)}-Z^*\|_2^2. \label{eq:gap_to_distance} \end{equation}
Rearranging yields
\begin{equation} \|Z^{(K)}-Z^*\|_2\le \sqrt{\frac{2\varepsilon_K}{\rho}}. \label{eq:distance_bound_final} \end{equation}
If the inner solver satisfies the standard accelerated objective rate $\varepsilon_K=O(K^{-2})$, then
\begin{equation} \|Z^{(K)}-Z^*\|_2=O\left(\frac{1}{\sqrt{\rho}K}\right). \label{eq:rate_final} \end{equation}
\end{proof}

\subsection{Proof of Zero-Block Recovery in ADMM}
\label{sec:kkt_margin_hard_zero}

\begin{proof}
For the proximal problem in~\eqref{eq:z_prox}, first-order optimality gives
\begin{equation} 0\in \rho(Z^*-Q)+\lambda\partial\Omega^2(Z^*). \label{eq:prox_kkt_margin_new} \end{equation}
Restricting~\eqref{eq:prox_kkt_margin_new} to block $u$ and considering $Z_u^*=0$ yields
\begin{equation} \rho Q_u\in\lambda\partial_u\Omega^2(0;Z^*_{-u}). \label{eq:zero_block_subgradient} \end{equation}
Thus, exact zero-block recovery is determined by whether $\rho Q_u$ lies inside the corresponding block-wise subgradient set. Define the distance to its zero/nonzero boundary as
\begin{equation} m_u:=\operatorname{dist}\!\left(\rho Q_u,\partial\!\left[\lambda\partial_u\Omega^2(0;Z^*_{-u})\right]\right). \label{eq:zero_block_margin} \end{equation}
If the perturbation induced by the approximate proximal solution is smaller than $m_u$, it cannot cross this boundary, so $Z^{(K)}$ and $Z^*$ have the same zero/nonzero status for block $u$ up to numerical tolerance. By Proposition~\ref{prop:admm_residual}, a sufficient condition is therefore
\begin{equation} \sqrt{\frac{2\varepsilon_K}{\rho}}<m_u. \label{eq:zero_recovery_sufficient} \end{equation}

For the hierarchical penalty, a useful local scale is
\begin{equation} \rho_c^{\mathrm{proxy}}=\frac{d_v}{\|f_v(W)^{p-1}\|_{\infty}},\qquad f_v(W)=\left(\sum_{u\in D(v)}\|W_u\|_2^p\right)^{1/p}. \label{eq:rho_c_proxy_new} \end{equation}
where $d_v=2^{|v|}$ and $p\in(1,2]$. This is a local heuristic scale rather than a universal critical value; the exact zero-block decision remains governed by~\eqref{eq:zero_block_subgradient}. Hence, once the finite-step residual falls below the local KKT margin, the approximate and exact proximal updates recover the same support up to numerical tolerance.
\end{proof}

% =============================================================================
% Appendix C: Proof of Proposition 3 (Active-Set KKT Ceiling)
% =============================================================================
\subsection{Proof of Proposition~\ref{prop:ceiling} (Active-Set KKT Ceiling)}
\label{app:prop3}

\begin{proof}
For each pair $(s,t)$, define
\begin{equation} g_{st}:=\nabla_W\mathcal{L}(W^*)_{st}. \label{eq:gst_def_prop3} \end{equation}
Using the triangle inequality and the gradient bound in~\eqref{eq:gradient_bound_prop3}, we have
\begin{equation} \mathrm{score}(v)=\frac{1}{d_v}\left\|\sum_{s,t}g_{st}\phi_v(X_{st})\right\|_2\le \frac{1}{d_v}\sum_{s,t}\|g_{st}\|_2|\phi_v(X_{st})|\le \frac{G}{d_v}\sum_{s,t}\phi_v(X_{st}). \label{eq:kkt_score_upper_bound} \end{equation}
Since $\phi_v(X_{st})$ is a product of independent Bernoulli variables,
\begin{equation} \mathbb{E}[\phi_v(X_{st})]=\mathbb{E}\left[\prod_{j\in v}x_j(s,t)\right]=\prod_{j\in v}\mathbb{E}[x_j(s,t)]=\prod_{j\in v}p_j\le p_{\max}^k. \label{eq:ncf_expectation_prop3} \end{equation}
Taking expectation in~\eqref{eq:kkt_score_upper_bound} gives
\begin{equation} \mathbb{E}[\mathrm{score}(v)]\le \frac{G}{d_v}\sum_{s,t}\mathbb{E}[\phi_v(X_{st})]\le \frac{G n_{\mathrm{eff}}p_{\max}^k}{2^k}=G n_{\mathrm{eff}}\left(\frac{p_{\max}}{2}\right)^k. \label{eq:kkt_decay_proof_final} \end{equation}
This proves~\eqref{eq:kkt_decay_rigorous}.

To obtain the threshold length, require
\begin{equation} G n_{\mathrm{eff}}\left(\frac{p_{\max}}{2}\right)^k\le \tau_{\mathrm{KKT}}. \label{eq:kstar_condition_proof} \end{equation}
Since $0<p_{\max}/2<1$, this condition is equivalent to
\begin{equation} k\ge \frac{\log(G n_{\mathrm{eff}}/\tau_{\mathrm{KKT}})}{\log(2/p_{\max})}. \label{eq:kstar_solve_proof} \end{equation}
Taking the ceiling gives the $k^*$ condition.

Finally, since $\mathrm{score}(v)\ge0$, Markov's inequality gives
\begin{equation} \Pr\left[\mathrm{score}(v)\ge \tau_{\mathrm{KKT}}+\epsilon\right]\le \frac{\mathbb{E}[\mathrm{score}(v)]}{\tau_{\mathrm{KKT}}+\epsilon}\le \frac{G n_{\mathrm{eff}}\left(p_{\max}/2\right)^k}{\tau_{\mathrm{KKT}}+\epsilon}. \label{eq:markov_proof_final} \end{equation}
This proves the exponential decay claim.
\end{proof}

\subsection{Proof of Corollary~\ref{cor:active_set_accessibility}}\label{app:active_set_accessibility}

\begin{proof}
The active-set rule can only add NCFs that are considered by the search and pass the KKT threshold. Any $u\in U^*$ with $|u|>D$ is excluded by the maximum-length constraint. Any $u\in U^*$ with $\mathbb{E}[\mathrm{score}(u)]\le\tau_{\mathrm{KKT}}$ is below the expected expansion threshold by Proposition~\ref{prop:ceiling}. Therefore, in expectation, the recoverable true support is contained in $U^*_{\mathrm{acc}}$. Since
\begin{equation} \mathrm{Recall}(\hat{S};U^*)=\frac{|\hat{S}\cap U^*|}{|U^*|}, \label{eq:recall_def_accessibility} \end{equation}
we obtain
\begin{equation} \mathbb{E}\left[\mathrm{Recall}(\hat{S};U^*)\right]=\frac{\mathbb{E}[|\hat{S}\cap U^*|]}{|U^*|}\le \frac{|U^*_{\mathrm{acc}}|}{|U^*|}. \label{eq:recall_bound_accessibility_proof} \end{equation}
\end{proof}

\section{Additional Diagnostics}\label{app}

\subsection{WGL Regularization-Scale Sensitivity}
\label{app:wgl_sensitivity}

We investigate the sensitivity of WGL to the regularization scale. Under the default continuation schedule ($\lambda:500\to50$), WGL collapses to an empty selected set with AUC $=0.500$ on all 10 civil-unrest datasets, whose positive-class fractions range from $0.5\%$ to $1.5\%$. In contrast, the flu dataset, with a positive-class fraction of $7.4\%$, does not fully collapse. This pattern suggests that the default WGL scale is particularly aggressive in the severely imbalanced setting, although the comparison does not establish class imbalance as the sole cause. The relationship between positive-class fraction and WGL behavior is further examined in Appendix~\ref{sec:imbalance_effect}.

To determine whether the failure reflects the WGL formulation itself or its regularization scale, we vary $\lambda$ while keeping the remaining training configuration fixed. Table~\ref{tab:wgl_lambda_sweep} reports a representative sweep on Venezuela (positive rate $0.7\%$). The default $500\to50$ continuation produces an empty support and chance-level AUC. Reducing the regularization scale restores nonzero supports and predictive performance: a fixed $\lambda=1$ selects seven NCFs and increases AUC from $0.500$ to $0.581$. Further decreasing $\lambda$ to $0.1$ increases the selected set to 13 NCFs but reduces AUC to $0.536$, showing that weaker regularization does not necessarily improve predictive performance.

These results identify regularization scale as an important factor in WGL behavior. The default collapse therefore should not be interpreted as an inherent inability of WGL to select predictive hierarchical features. Based on this diagnostic, we use WGL($\lambda=1$) as the non-collapsing WGL configuration in the main predictive comparison.

\begin{table}[!ht]
\centering
\caption{%
Sensitivity of WGL to the regularization scale on Venezuela.
``Direction'' indicates either high-to-low continuation from $\lambda_{\mathrm{start}}$ to $\lambda_{\mathrm{end}}$ or a single fixed-$\lambda$ run.
``sel@0.40'' denotes the number of NCFs selected at the active-set threshold $\tau=0.40$.
}
\label{tab:wgl_lambda_sweep}
\small
\setlength{\tabcolsep}{5pt}
\begin{tabular}{ccccc}
\toprule
$\lambda_{\mathrm{start}}$ & $\lambda_{\mathrm{end}}$ & Direction & AUC & sel@0.40 \\
\midrule
500 & 50  & high-to-low & 0.500 & 0 \\
50  & 5   & high-to-low & 0.572 & 1 \\
5   & 0.5 & high-to-low & 0.540 & 9 \\
\textbf{1} & \textbf{1} & single & \textbf{0.581} & \textbf{7} \\
0.1 & 0.1 & single & 0.536 & 13 \\
\bottomrule
\end{tabular}
\end{table}

\subsection{Paired-Seed Analysis of the ADMM Proximal Update}\label{sec:admm_paired_analysis}
\begin{figure}[!ht]

\centering
\includegraphics[width=0.55\linewidth]{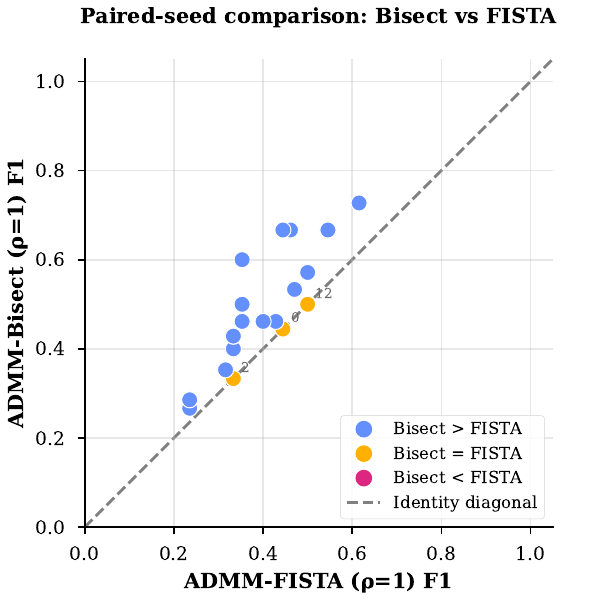}
\caption{Paired-seed F1 comparison between ADMM-Bisect and ADMM-FISTA at $\rho=1$. Each point represents the same random seed evaluated with the two proximal solvers; points above the identity diagonal favor ADMM-Bisect. Under this diagnostic condition, ADMM-Bisect matches or improves ADMM-FISTA on every evaluated seed, showing that the aggregate improvement is consistent across paired runs rather than driven by a small number of seeds.}
\label{fig:admm_paired}
\par
\end{figure}

To examine whether the aggregate improvement from bisection is consistent across runs, Figure~\ref{fig:admm_paired} compares ADMM-Bisect and ADMM-FISTA seed by seed at $\rho=1$. ADMM-Bisect matches or improves ADMM-FISTA on every paired seed in this diagnostic, with most points lying above the identity diagonal. Together with the Holm-survived aggregate improvement ($\Delta\mathrm{F1}=+0.088$, $p_{\mathrm{adj}}=2.60\times10^{-3}$), this result provides consistent empirical support for replacing the finite-step FISTA proximal update with bisection.

% Finally, Figures~\ref{fig:convergence} and~\ref{fig:wgl_failure} separate benign equivalence from genuine failure. The convergence traces confirm that the earlier VQF discrepancy was a hyperparameter mismatch, not a new optimizer. The WGL diagnostic shows the opposite: WGL's collapse is a real scale-dependent failure mode, and the adaptive schedule can make the collapse worse by reinforcing early large groups while suppressing weaker true groups.

% \begin{figure}[!htbp]
% 
%   \centering
%   \includegraphics[width=\linewidth]{figures/fig6_convergence.pdf}
%   \caption{%
%     Convergence diagnostics for MM, VQF, ADMM, and WGL variants.
%     MM and VQF produce identical iterates under matched settings, while the other solvers optimize different split or reduced objectives and should not be compared by raw objective scale alone.
%   }
%   \label{fig:convergence}
% \par
% \end{figure}

% \begin{figure}[!htbp]
% 
%   \centering
%   \includegraphics[width=\linewidth]{figures/fig7_wgl_failure.pdf}
%   \caption{%
%     WGL failure modes in the synthetic diagnostics.
%     Plain continuation is bimodal across seeds, and the adaptive schedule collapses to very small selected supports.
%     This negative result motivates the real-data $\lambda$-scale analysis in Section~\ref{app:wgl_sensitivity}.
%   }
%   \label{fig:wgl_failure}
% \par
% \end{figure}

\section{Active-Set Mechanism Analysis}
\label{sec:active_set_analysis}

All HTNL optimization methods studied in this paper share the same outer active-set procedure. At each iteration, inactive candidate NCFs are scored using the KKT criterion in Equation~\ref{eq:kkt_score_background}, the top-$K$ candidates are added to the active set, and the restricted optimization problem is re-solved. The final selected set $\hat{S}$ is determined by the threshold $\tau$. Because this outer procedure is shared across methods, differences in selected NCFs arise from the inner optimization and its effect on subsequent KKT scores.

\subsection{Per-Method Active-Set Behavior}
\label{sec:active_set_per_method}

The optimization methods exhibit distinct selection behavior. \textbf{MM/VQF} produces compact and stable supports: the two methods select identical NCF sets on all real-data benchmarks, with Jaccard similarity $1.000$, consistent with Proposition~\ref{prop:equiv}. At $\tau=0.40$, they typically select about $4$--$5$ NCFs on the LatAm datasets. \textbf{ADMM-Bisect} selects broader supports. Although bisection removes the finite-step FISTA proximal error in the ADMM $Z$-update, its selected sets remain larger and have low overlap with MM/VQF, showing that a more accurate proximal update does not recover the same compact support. \textbf{WGL} is sensitive to the regularization scale: the default continuation path $\lambda_{\mathrm{init}}=500\to\lambda_{\mathrm{target}}=50$ produces empty supports on the LatAm datasets, whereas $\lambda=1$ restores nonzero supports. Under this setting, WGL is closer to ADMM-Bisect than to MM/VQF, with Jaccard similarities of $0.780$ and $0.351$, respectively. \textbf{WGL-Adaptive} instead produces very small selected sets, indicating that its adaptive weighting can strongly suppress weaker candidates.

\subsection{Comparative Summary}
\label{sec:active_set_summary}

Table~\ref{tab:active_set_summary} summarizes the selected-support behavior on the LatAm datasets. MM/VQF produces the most compact supports, selecting approximately $4$--$5$ NCFs while achieving the largest mean AUC improvement. ADMM-Bisect and WGL($\lambda=1$) select broader supports of approximately 10 NCFs and show substantially lower overlap with MM/VQF. In contrast, the default WGL continuation produces empty supports. These results show that different inner optimization methods can lead to substantially different selected sets even when they share the same outer active-set procedure.

\begin{table}[!ht]
\centering
\caption{Selected-support behavior on the LatAm civil-unrest datasets. $|\hat{S}|$ denotes the selected set size at $\tau=0.40$, Jaccard similarity is measured against MM/VQF, and $\Delta$AUC is measured against class-weighted Logistic Regression.}
\label{tab:active_set_summary}
\small
\setlength{\tabcolsep}{7pt}
\begin{tabular}{lccc}
\toprule
\textbf{Method} & \textbf{Jac. vs Orig./VQF} & $\boldsymbol{|\hat{S}|}$ & $\boldsymbol{\Delta}$\textbf{AUC} \\
\midrule
MM/VQF       & $1.000$ & ${\approx}4$--$5$ & $+0.074^{***}$ \\
ADMM-Bisect        & $0.277$ & ${\approx}10$     & $+0.034^{*}$ \\
WGL($\lambda=1$)   & $0.351$ & ${\approx}10$     & $+0.044^{**}$ \\
WGL-Cont (default) & $0.000$ & $0$               & --- \\
\bottomrule
\end{tabular}
\end{table}
\subsection{Active-Set Depth Ceiling}
\label{sec:active_set_ceiling}

The shared active-set rule imposes a structural limitation. Proposition~\ref{prop:ceiling} shows that the KKT score for a length-$k$ NCF decays as $O((p_{\max}/2)^k)$, making sufficiently long NCFs increasingly unlikely to enter the active set for fixed $n_{\mathrm{eff}}$ and $\tau_{\mathrm{KKT}}>0$. Thus, improving the inner optimization method can change support size, predictive performance, and runtime, but does not directly remove the KKT-based candidate-admission limit.

Table~\ref{tab:active_set_sensitivity} provides empirical evidence for this limitation. Across expanded active-set configurations, VQF remains at an F1 plateau of $0.686$, while ADMM-Bisect varies only from $0.674$ to $0.685$. Increasing the active-set budget therefore does not overcome the observed ceiling, consistent with Proposition~\ref{prop:ceiling}. Overcoming this limitation requires modifying the candidate-expansion mechanism rather than only the restricted inner optimization.

\begin{table}[!ht]
\centering
\caption{Active-set sensitivity on the TKDE-faithful synthetic generator (Dataset 2). $m_0$ denotes the initial active-set size; the default configuration uses $m_0=2$ and adds five candidates per iteration. Entries report mean F1.}
\label{tab:active_set_sensitivity}
\small
\setlength{\tabcolsep}{8pt}
\begin{tabular}{lcccc}
\toprule
\textbf{Method} & \textbf{Default ($m_0=2$, add=5)} & \textbf{$m_0=3$} & \textbf{$m_0=4$} & \textbf{$m_0=5$} \\
\midrule
MM/VQF & $0.590$ & $0.686$ & $0.686$ & $0.686$ \\
ADMM-Bisect & $0.674$ & $0.684$ & $0.685$ & $0.685$ \\
\bottomrule
\end{tabular}
\end{table}

\subsection{Selected-Support Differences Across optimization methods}
\label{sec:realdata_jaccard}

We compare the NCF sets selected by different optimization methods. Table~\ref{tab:realdata_jaccard_v3} reports the mean Jaccard similarity at $\tau=0.40$ across the 11 real-data benchmarks.

\begin{table}[!ht]
\centering
\small
\setlength{\tabcolsep}{4pt}
\caption{Mean Jaccard similarity between selected NCF sets across the 11 real-data benchmarks at $\tau=0.40$. MM and VQF have Jaccard similarity $1.000$, consistent with Proposition~\ref{prop:equiv}.}
\label{tab:realdata_jaccard_v3}
\begin{tabular}{lccccc}
\toprule
 & \textbf{Orig} & \textbf{VQF} & \textbf{ADMM} & \textbf{WGL-def.} & \textbf{WGL($\lambda=1$)} \\
\midrule
\textbf{MM}         & \textbf{1.000} & \textbf{1.000} & $0.277$ & $0.045$ & $0.351$ \\
\textbf{VQF}              & \textbf{1.000} & \textbf{1.000} & $0.277$ & $0.045$ & $0.351$ \\
\textbf{ADMM-Bisect}      & $0.277$ & $0.277$ & \textbf{1.000} & $0.000$ & $0.780$ \\
\textbf{WGL-Cont (def.)}  & $0.045$ & $0.045$ & $0.000$ & \textbf{1.000} & $0.000$ \\
\textbf{WGL($\lambda=1$)} & $0.351$ & $0.351$ & $0.780$ & $0.000$ & \textbf{1.000} \\
\bottomrule
\end{tabular}
\par
\end{table}

MM and VQF select identical supports, as predicted by Proposition~\ref{prop:equiv}. ADMM-Bisect has substantially lower agreement with MM/VQF ($0.277$), showing that correcting the proximal update does not make ADMM select the same compact support. After regularization-scale correction, WGL($\lambda=1$) is more similar to ADMM-Bisect than to MM/VQF, with Jaccard similarities of $0.780$ and $0.351$, respectively. Thus, ADMM-Bisect and calibrated WGL tend to select similar, broader NCF sets.

\subsection{Selection Size and Outer Iterations}
\label{sec:granularity_niter}

\begin{figure}[!ht]

\centering
\begin{subfigure}[t]{0.48\linewidth}
\centering
\includegraphics[width=\linewidth]{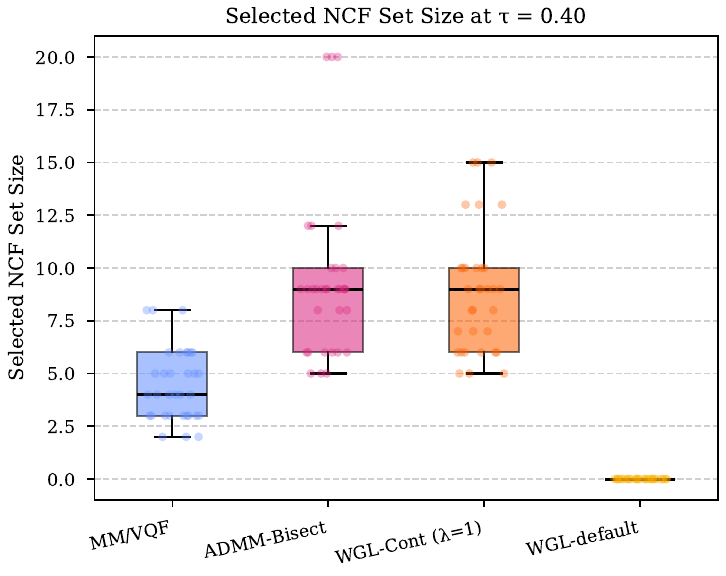}
\caption{Selected NCF set size at $\tau=0.40$.}
\label{fig:sel_size_box}
\end{subfigure}
\hfill
\begin{subfigure}[t]{0.48\linewidth}
\centering
\includegraphics[width=\linewidth]{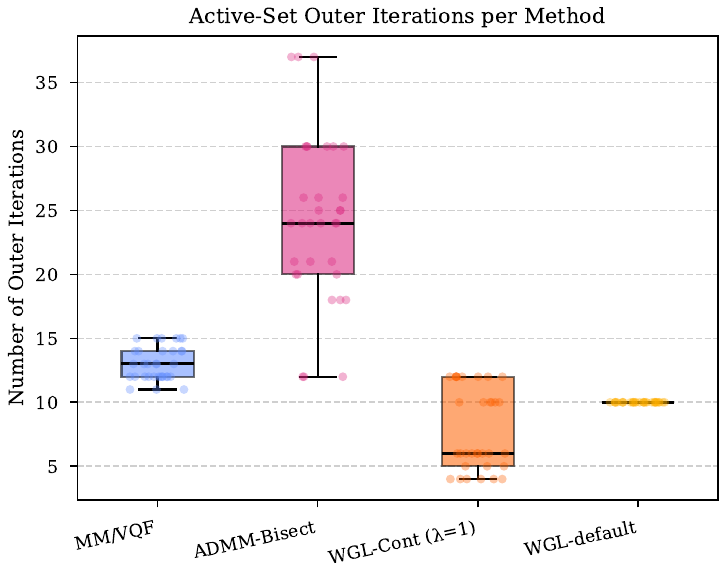}
\caption{Number of outer active-set iterations.}
\label{fig:niter_box}
\end{subfigure}
\caption{Selection size and outer active-set iterations across optimization methods. Each point represents one dataset--seed pair. MM/VQF produces more compact selected sets, whereas ADMM-Bisect and WGL($\lambda=1$) retain broader supports. ADMM-Bisect requires more outer iterations but remains computationally efficient in the tested implementation.}
\label{fig:sel_size_niter}
\par
\end{figure}

Figure~\ref{fig:sel_size_niter} compares selected-set size and the number of outer active-set iterations. MM/VQF produces the most compact supports, with a median of four selected NCFs and a standard deviation of $1.7$. ADMM-Bisect and WGL($\lambda=1$) both have a median selected-set size of nine, with standard deviations of $3.9$ and $2.9$, respectively. Thus, the broader supports of ADMM-Bisect and calibrated WGL are observed consistently across datasets and seeds. Default WGL instead produces empty supports on the 10 LatAm benchmarks.

The number of outer iterations shows a different pattern. MM/VQF requires approximately 13 iterations on average, compared with approximately 24 for ADMM-Bisect and 10 for WGL($\lambda=1$). Despite requiring more outer iterations, ADMM-Bisect remains faster in wall-clock time because its tested implementation reduces the cost of each iteration through KKT pruning. This distinction shows that the number of outer iterations alone does not determine total runtime.

\section{Additional Real-Data Results}
\label{app:realdata_details}

\subsection{\texorpdfstring{VQF $\equiv$ MM}{VQF equivalent to MM} on Real Data}
\label{sec:vqf_confirmed}

Proposition~\ref{prop:equiv} shows that VQF and MM perform the same weighted-ridge updates under matched auxiliary updates. This equivalence is also observed empirically on all 11 real-data benchmarks: VQF and MM select identical NCF sets, with Jaccard similarity $1.000$ across all evaluated runs. We therefore report the two methods jointly as MM/VQF in the remaining real-data analyses.

\subsection{Real-Data Selection and Predictive Performance}
\label{sec:realdata}

Although MM/VQF produces identical selected sets, ADMM-Bisect exhibits substantially different selection behavior. As shown in Table~\ref{tab:realdata_countries}, ADMM-Bisect selects $62$--$97$ NCFs on the five TKDE-replication civil-unrest datasets, compared with $13$--$25$ for MM/VQF. The mean selected-support size is $80.8$ for ADMM-Bisect and $18.6$ for MM/VQF. Thus, replacing the finite-step FISTA proximal update with bisection does not make the ADMM solution recover the more compact MM/VQF support. On the same five datasets, ADMM-Bisect also requires less runtime, averaging $5.6$ seconds per country compared with $25.5$ seconds for MM/VQF.

\begin{figure}[!ht]
\centering
\includegraphics[width=\linewidth]{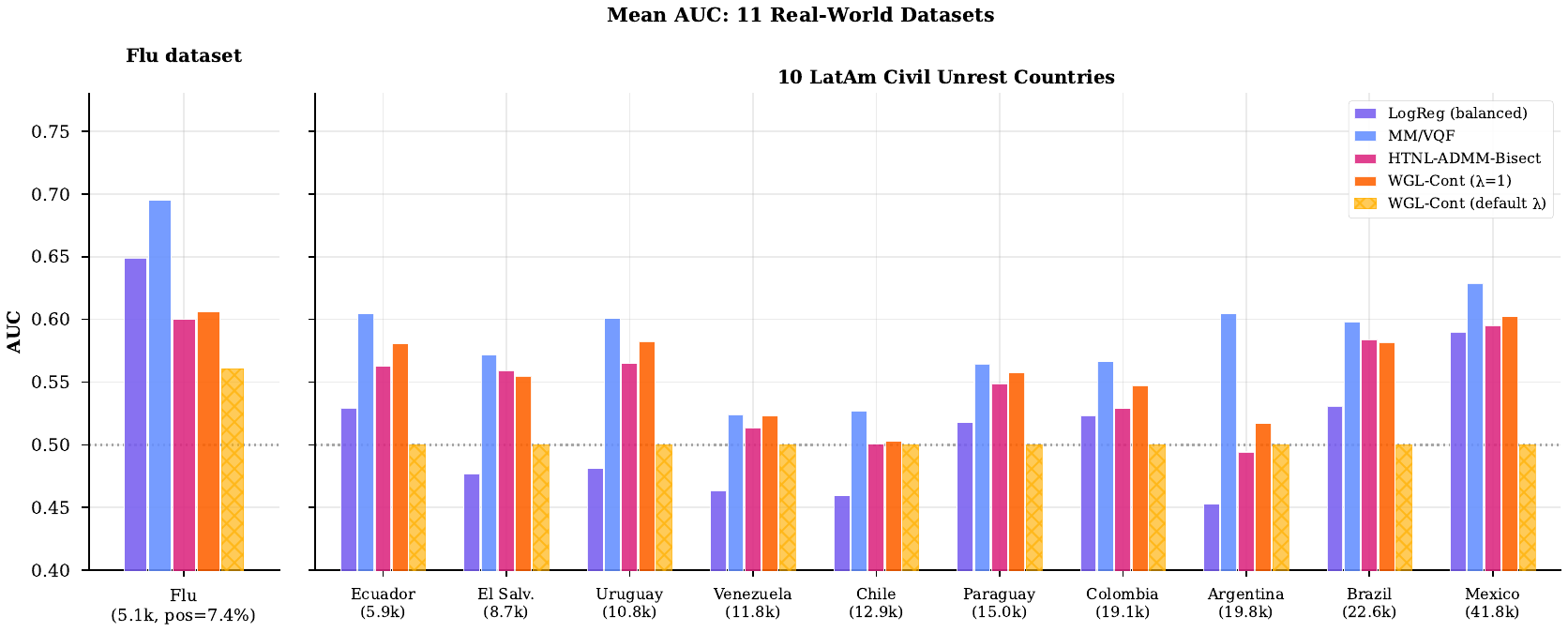}
\caption{Predictive AUC across the 11 real-data benchmarks. MM/VQF provides the largest mean improvement over class-weighted Logistic Regression. ADMM-Bisect and WGL($\lambda=1$) also improve mean AUC but retain broader selected supports, while the default WGL continuation produces empty models on the severely imbalanced civil-unrest datasets.}
\label{fig:realdata_auc}
\end{figure}

Figure~\ref{fig:realdata_auc} summarizes predictive performance across all 11 datasets. MM/VQF improves over class-weighted Logistic Regression on every dataset, while ADMM-Bisect and WGL($\lambda=1$) also yield positive mean AUC gains. The broader supports selected by ADMM-Bisect and WGL($\lambda=1$) do not translate into higher AUC than the more compact MM/VQF solutions. In contrast, the default WGL continuation schedule produces empty models on the 10 severely imbalanced civil-unrest datasets, showing its sensitivity to the regularization scale. We use AUC as the primary predictive metric because these datasets are highly imbalanced.

\begin{table}[!ht]
\centering
\caption{Selected-support sizes on five civil-unrest datasets. $n_{\mathrm{train}}$ and pos\% denote the number of training instances and the positive-class fraction, respectively. Support size is measured at $\tau=0.40$. MM and VQF are identical under matched updates and are therefore reported jointly. ADMM-Bisect uses $\rho=1$.}
\label{tab:realdata_countries}
\small
\setlength{\tabcolsep}{5pt}
\begin{tabular}{lrrcc}
\toprule
\textbf{Country} & $n_{\text{train}}$ & \textbf{pos\%} & \textbf{MM/VQF} & \textbf{ADMM-Bisect} \\
\midrule
Uruguay   & 10,788 & 0.8\% & 13 & 76 \\
Venezuela & 11,832 & 0.7\% & 15 & 62 \\
Paraguay  & 14,964 & 1.5\% & 22 & 86 \\
Colombia  & 19,140 & 0.5\% & 25 & 97 \\
Argentina & 19,836 & 0.5\% & 18 & 83 \\
\midrule
\textbf{Mean} & --- & --- & \textbf{18.6} & \textbf{80.8} \\
\bottomrule
\end{tabular}
\end{table}

\subsection{Runtime and Efficiency Analysis}
\label{sec:runtime}

Figure~\ref{fig:runtime} reports the mean wall-clock time of each optimization method on the 10 LatAm civil-unrest datasets and the flu dataset. All runs use a single CPU process on the same machine.

On the 10 LatAm datasets, MM and VQF have nearly identical runtimes, averaging approximately $25$\,s per dataset. ADMM-Bisect averages approximately $6$\,s, corresponding to a $3$--$4\times$ speedup. The tested implementation combines bisection with KKT pruning, which removes candidates with scores below $\tau_{\mathrm{prune}}=0.10$ before the inner ADMM solve and reduces the size of the resulting subproblem (Section~\ref{sec:admm_bisect_contribution} and Appendix~\ref{app:kkt_speedup}).

WGL-Cont has greater runtime variation. It is fast when the continuation procedure terminates early with an empty support, but becomes slower when multiple $\lambda$ stages are evaluated. On Mexico, the largest LatAm dataset ($n=41{,}760$), WGL-Cont takes $140$\,s, compared with approximately $25$\,s for MM.

On the flu dataset (pos$\%=7.4\%$), ADMM-Bisect takes approximately $18$\,s and MM approximately $22$\,s. WGL($\lambda=1$) is fastest at approximately $9$\,s because the fixed-$\lambda$ configuration avoids the multi-stage continuation procedure.

Overall, ADMM-Bisect remains computationally efficient despite using an accurate proximal update. The scalar bisection solver requires only $O(\log(1/\varepsilon))$ iterations, while KKT pruning further reduces the number of blocks processed by the inner solver.

\subsection{Per-Dataset AUC}
\label{sec:perdataset}

\begin{figure}[!ht]
\centering
\begin{subfigure}[t]{0.48\linewidth}
\centering
\includegraphics[width=\linewidth,height=0.34\textheight,keepaspectratio]{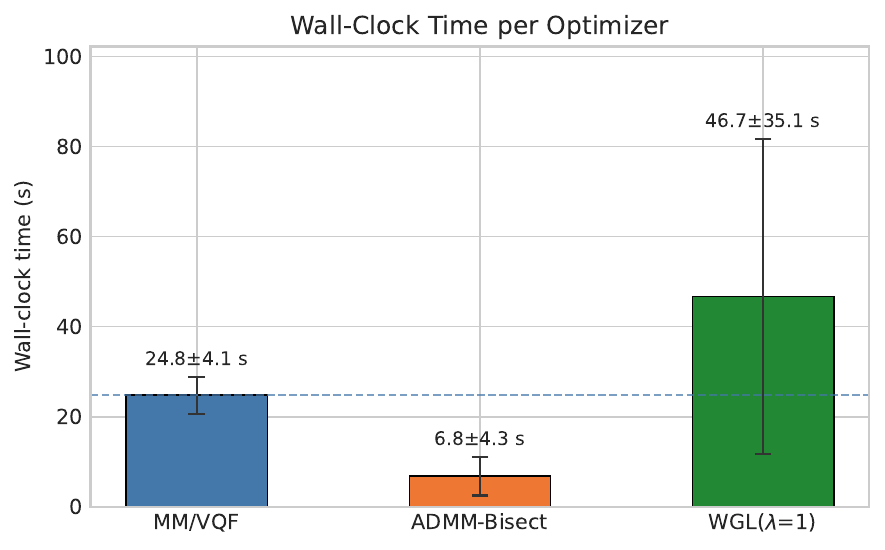}
\caption{Mean wall-clock time. Error bars show $\pm1$ standard deviation across datasets.}
\label{fig:runtime}
\end{subfigure}
\hfill
\begin{subfigure}[t]{0.48\linewidth}
\centering
\includegraphics[width=\linewidth,height=0.25\textheight,keepaspectratio]{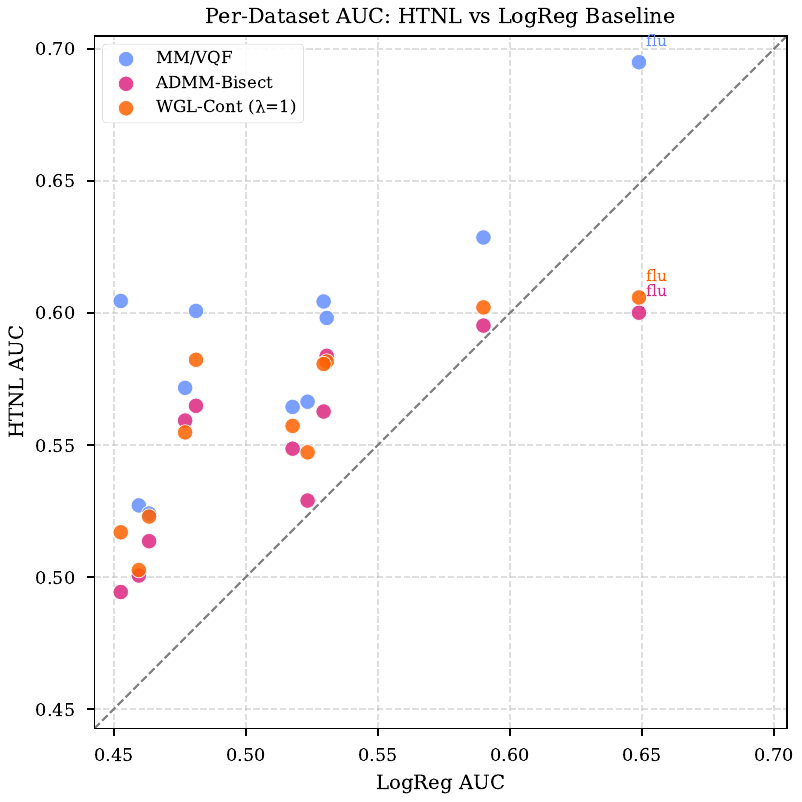}
\caption{Per-dataset AUC relative to Logistic Regression. Points above the diagonal indicate higher AUC than the baseline.}
\label{fig:auc_scatter}
\end{subfigure}
\caption{Runtime and predictive performance across the 10 LatAm civil-unrest datasets and the flu dataset. The left panel compares wall-clock time, and the right panel compares per-dataset AUC with class-weighted Logistic Regression.}
\label{fig:runtime_auc}
\end{figure}

Figure~\ref{fig:auc_scatter} shows the per-dataset AUC results. MM/VQF achieves higher AUC than ADMM-Bisect and WGL($\lambda=1$) on all 10 LatAm datasets, with particularly large improvements over Logistic Regression on Argentina ($+0.152$) and Uruguay ($+0.120$). ADMM-Bisect and WGL($\lambda=1$) generally remain closer to the Logistic Regression baseline and also select broader NCF sets. Thus, the additional features retained by these methods do not produce higher AUC on the LatAm benchmarks.

The flu dataset differs from the LatAm datasets in both positive-class fraction and performance pattern. Its positive rate is $7.4\%$, compared with $0.48$--$1.52\%$ for the LatAm datasets, and the differences among the HTNL methods are smaller. Because only one dataset lies outside the severely imbalanced LatAm range, we treat this pattern as descriptive rather than evidence for a specific class-imbalance threshold.

\subsection{Performance across Class-Imbalance Levels}
\label{sec:imbalance_effect}

\begin{figure}[!ht]
\centering
\includegraphics[width=0.78\linewidth]{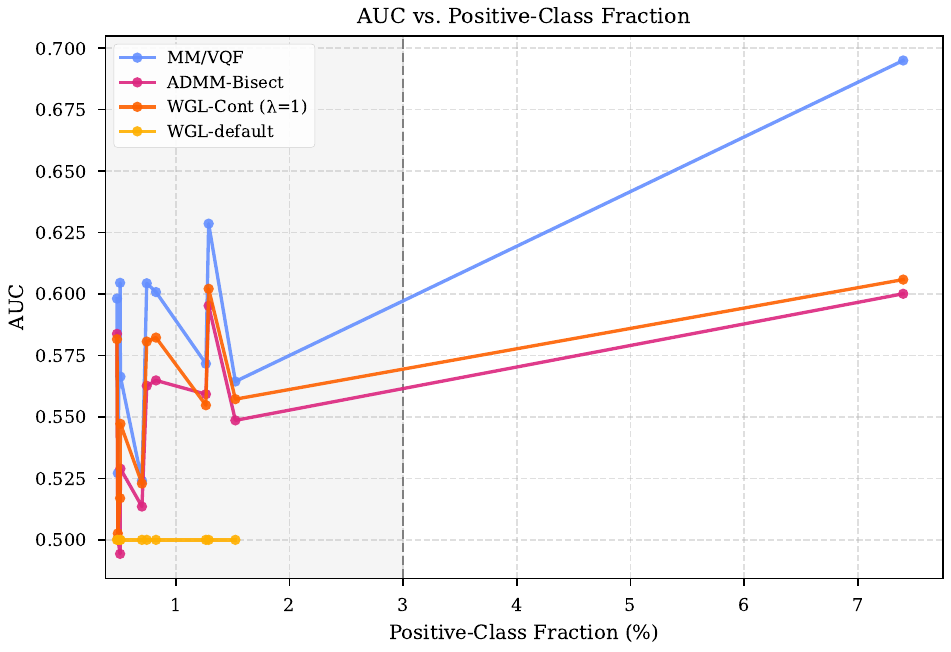}
\caption{AUC as a function of the positive-class fraction. The 10 LatAm datasets have positive rates between $0.48\%$ and $1.52\%$, whereas the flu dataset has a positive rate of $7.4\%$. Default WGL produces empty supports with AUC $=0.500$ on all 10 LatAm datasets, while this failure is not observed on the flu dataset.}
\label{fig:posrate_auc}
\end{figure}

Figure~\ref{fig:posrate_auc} examines performance as a function of the positive-class fraction. On all 10 severely imbalanced LatAm datasets, default WGL produces empty supports and AUC $=0.500$, whereas MM/VQF, ADMM-Bisect, and WGL($\lambda=1$) produce nonzero solutions. MM/VQF achieves the highest AUC among the evaluated HTNL methods on each of these datasets. This pattern shows that the default WGL regularization scale is poorly matched to the LatAm setting, while reducing $\lambda$ restores nonzero solutions.

The flu dataset, with a positive rate of $7.4\%$, does not show the same default-WGL failure pattern, and the AUC differences among the non-collapsing methods are smaller. This contrast suggests an association between the data scale, class imbalance, and WGL sensitivity, but the current benchmarks do not identify a specific positive-rate boundary or establish class imbalance as the sole cause of the collapse.

\section{Systematic Optimization Method Comparison}
\label{sec:sys_comparison}

The four HTNL optimization families studied in this paper differ along several technical dimensions, including objective formulation, inner optimization routine, convergence behavior, per-iteration computational cost, active-set selection behavior, robustness to class imbalance, and hyperparameter requirements. This section systematically compares these differences, connects them to the mechanisms identified in our analysis, and summarizes practical guidelines for choosing an optimization method.

\subsection{Multi-Dimensional Overview}
\label{sec:comparison_table}

Table~\ref{tab:optimizer_comparison} compares the four optimization families across eight dimensions. MM and VQF use the same variational quadratic formulation and produce identical updates, whereas ADMM-Bisect directly handles the nonsmooth objective through ADMM and WGL converts the hierarchical penalty into a sequence of weighted group-lasso problems. These different optimization strategies lead to distinct convergence, computational, and feature-selection behavior.

\begin{table}[!ht]
\centering
\caption{Multi-dimensional comparison of the four HTNL optimization families. ``Sparse'' denotes the sparse-selection regime (${\approx}4$--$5$ selected NCFs), ``broad'' denotes the broader-selection regime (${\approx}10$ selected NCFs), and ``collapse'' denotes an empty or nearly empty selected set. ``Ridge'' refers to the weighted ridge subproblem obtained after fixing the variational auxiliary variables, and ``FISTA'' refers to the accelerated proximal-gradient solver used for the weighted group-lasso subproblem. The parameter $\rho\in\mathbb{R}_+$ is the ADMM augmented-Lagrangian penalty, while the $\lambda$ schedule denotes the continuation path used by WGL.}
\label{tab:optimizer_comparison}
\scriptsize
\setlength{\tabcolsep}{2pt}
\renewcommand{\arraystretch}{1.08}
\begin{threeparttable}
\resizebox{\linewidth}{!}{%
\begin{tabular}{>{\raggedright\arraybackslash}p{2.1cm}>{\raggedright\arraybackslash}p{2.55cm}>{\raggedright\arraybackslash}p{2.45cm}>{\raggedright\arraybackslash}p{2.75cm}>{\raggedright\arraybackslash}p{2.85cm}}
\toprule
\textbf{Dimension} & \textbf{MM} & \textbf{VQF} & \textbf{ADMM-Bisect} & \textbf{WGL-Cont} \\
\midrule
\textbf{Formulation basis} & Variational quadratic formulation & Same variational quadratic formulation as MM & ADMM decomposition of the original nonsmooth objective & Reformulation of the outer $\ell_p$ hierarchy as weighted group lasso \\
\addlinespace
\textbf{Inner solver} & Weighted ridge subproblem & Same weighted ridge subproblem as MM & L-BFGS for the smooth step plus exact bisection for the nonsmooth proximal step & FISTA for the weighted group-lasso subproblem \\
\addlinespace
\textbf{Convergence behavior} & Linear convergence for the strongly convex ridge subproblem & Same as MM & $O(1/K)$ ADMM residual convergence & $O(1/K^2)$ convergence within each fixed-$\lambda$ FISTA stage; no recovery guarantee after support collapse \\
\addlinespace
\textbf{Per-iteration cost} & $O(|U|\cdot\text{ridge})$ & $O(|U|\cdot\text{ridge})$ & $O(|U|\cdot|D(v)|\cdot\log(1/\varepsilon))$ & $O(|U|\cdot G\cdot\text{FISTA})$ \\
\addlinespace
\textbf{Selection regime} & Sparse: ${\approx}4$--$5$ NCFs & Identical to MM & Broad: ${\approx}10$ NCFs & Broad when $\lambda{=}1$; collapse under the default continuation schedule \\
\addlinespace
\textbf{Class-imbalance robustness} & Robust in the tested rare-event datasets & Robust in the tested rare-event datasets & Robust in the tested rare-event datasets & Scale-sensitive; the default schedule collapses on all 10 severely imbalanced civil-unrest datasets \\
\addlinespace
\textbf{Extra hyperparameters} & None beyond shared $C$ and $\tau$ & None beyond shared $C$ and $\tau$ & $\rho$, the augmented-Lagrangian penalty & $\lambda_{\mathrm{init}}$, $\lambda_{\mathrm{target}}$, and the number of continuation steps \\
\addlinespace
\textbf{When to use} & Sparse, interpretable selection with minimal tuning & Same as MM, since the updates are algebraically equivalent & Broader support recovery with exact proximal computation and stable AUC across tested $\rho$ values & Broader support or regularization-path analysis; requires $\lambda$ calibration on severely imbalanced data \\
\bottomrule
\end{tabular}%
}
\end{threeparttable}
\end{table}

% --------------------------------------------------------------------------
\subsection{Formulation Equivalences and Distinctions}
\label{sec:comparison_formulation}

The four methods can be grouped into three formulations. MM and VQF use the same variational quadratic formulation; ADMM-Bisect directly optimizes the original nonsmooth objective using ADMM; and WGL rewrites the hierarchical penalty as a weighted one-level group-lasso problem. These differences lead to distinct optimization behavior, failure modes, and selected feature sets.

\textbf{MM/VQF: equivalent weighted-ridge updates.} MM and VQF introduce auxiliary variables $(\gamma,\lambda,\mu)$ to express the two-level HTNL regularizer as a weighted quadratic objective; see Lemma~\ref{lemma:1}. Under matched initialization and exact auxiliary updates, Proposition~\ref{prop:equiv} shows that they solve the same weighted-ridge subproblem at every iteration. We therefore treat them as one method family, VQF. The weighted-ridge update strongly shrinks marginally active weights toward zero, leading VQF to select compact supports in our experiments. This behavior makes it suitable when sparse, interpretable feature selection and minimal tuning are preferred over recovering a broader set of features.

\textbf{ADMM-Bisect: direct nonsmooth optimization with exact proximal updates.} ADMM-Bisect retains the original two-level group-norm regularizer and uses ADMM to separate the smooth loss from the nonsmooth regularization; see Equation~\ref{eq:admm-constrained-problem}. The $W$-update optimizes the smooth loss with a quadratic agreement penalty, while the $Z$-update applies the proximal operator of the local two-level regularizer. ADMM-Bisect replaces the truncated FISTA approximation in the $Z$-update with a bisection solver. Proposition~\ref{prop:admm_residual} shows that finite-step proximal error can leave blocks near the zero/nonzero boundary incorrectly nonzero, whereas bisection solves the proximal equations to numerical tolerance. ADMM-Bisect nevertheless selects broader supports than VQF because its intermediate supports can increase the KKT scores of related inactive candidates. It is therefore useful when accurate proximal computation and broader feature recovery are desired.

\textbf{WGL: weighted group-lasso reformulation.} WGL uses the dual representation of the outer $\ell_p$ norm to rewrite the two-level hierarchy as a weighted one-level group-lasso problem; see Equation~\ref{eq:lp-dual-representation}. For fixed weights, the resulting group-lasso problem can be solved efficiently, after which the weights are updated to reflect the hierarchical penalty. The main practical limitation is sensitivity to the regularization scale. When the initial $\lambda$ is too large, the first continuation stage can shrink all groups to zero, making subsequent recovery difficult. In our experiments, the default schedule $\lambda_{\mathrm{init}}=500\to\lambda_{\mathrm{target}}=50$ collapses on all 10 severely imbalanced civil-unrest datasets, whereas $\lambda=1$ restores nonzero supports and predictive performance. WGL therefore requires careful $\lambda$ calibration on severely imbalanced data.

% --------------------------------------------------------------------------
\subsection{Convergence and Computational Cost}
\label{sec:comparison_convergence}

The optimization families differ substantially in both convergence behavior and wall-clock cost. Faster convergence of an individual inner problem does not necessarily lead to lower total runtime, because the overall cost also depends on the number of active blocks, candidate screening, and repeated active-set updates.

\paragraph{VQF: fast inner convergence but moderate overall cost.} Each outer iteration of VQF solves a strongly convex weighted ridge subproblem, for which standard first-order solvers achieve linear convergence under the usual smoothness and strong-convexity conditions. The per-iteration cost is $O(|U|\cdot P^2)$. Across the 11 benchmarks, VQF takes $26.7\,\mathrm{s}$ on average, making it the slowest non-collapsing variant in our comparison. Much of this cost comes from repeatedly solving the ridge problem over the full active set, including blocks whose weights remain small.

\paragraph{ADMM-Bisect: slower theoretical rate but lower total runtime.} For fixed $\rho$, standard ADMM has an $O(1/K)$ convergence rate under appropriate convexity assumptions~\citep{Boyd2011}. Despite this slower theoretical rate, ADMM-Bisect requires only $19.8\,\mathrm{s}$ on average, compared with $26.7\,\mathrm{s}$ for VQF, corresponding to a $1.3{\times}$ overall speedup on the same hardware. We further evaluate a computational variant that combines KKT pruning with the bisection proximal solver. On the synthetic diagnostic, this configuration preserves essentially the same F1 as baseline ADMM while reducing wall-clock time from approximately $15\,\mathrm{s}$ to $4\,\mathrm{s}$, corresponding to a $3.9{\times}$ speedup (Appendix~\ref{app:kkt_speedup}). Thus, practical efficiency depends not only on the convergence rate of the inner solver but also on reducing unnecessary computation during optimization.

\paragraph{WGL: fast inner optimization but sensitive continuation.} For fixed $\lambda$, the FISTA solver used by WGL achieves an $O(1/K^2)$ objective convergence rate under standard assumptions~\citep{Beck2009}. This guarantee applies to each fixed-$\lambda$ subproblem rather than to the complete continuation procedure. In our experiments, a large initial $\lambda$ can produce an empty support on the severely imbalanced civil-unrest datasets, whereas a calibrated regularization scale restores non-degenerate solutions. WGL therefore combines efficient fixed-$\lambda$ optimization with greater sensitivity to the regularization schedule and requires explicit $\lambda$ calibration in the severely imbalanced setting.

% --------------------------------------------------------------------------
\subsection{Selection Behavior and Support Recovery}
\label{sec:comparison_selection}

The optimization affects not only predictive performance but also the size and composition of the selected NCF set. Among the non-collapsing methods, VQF consistently selects compact supports, whereas ADMM-Bisect and WGL($\lambda{=}1$) retain substantially broader supports. The default WGL configuration represents a separate failure mode in which the selected support becomes empty.

\paragraph{Compact selection: VQF.} VQF produces the smallest selected sets, with approximately $4$--$5$ NCFs per dataset at $\tau{=}0.40$. This compact selection does not reduce predictive performance: VQF achieves the largest improvement over Logistic Regression ($\Delta\mathrm{AUC}{=}{+}0.074$, $p{=}4{\times}10^{-5}$). MM and VQF also select identical supports on all 11 datasets, with Jaccard similarity $1.000$, consistent with their equivalent weighted-ridge updates. VQF is therefore well suited when a small and interpretable set of selected indicators is desired.

\paragraph{Broader selection: ADMM-Bisect and WGL($\lambda{=}1$).} ADMM-Bisect and WGL($\lambda{=}1$) select approximately 10 NCFs on average, roughly twice the support size of VQF. Their selected sets are also similar to each other, with Jaccard similarity $0.780$, the largest off-diagonal value in the agreement matrix. Thus, after correcting the WGL regularization scale, both methods exhibit a similar broad-selection pattern.

For ADMM-Bisect, the broader support cannot be explained solely by an inaccurate proximal update. Replacing truncated FISTA with bisection removes the proximal residual described in Proposition~\ref{prop:admm_residual}, but the selected support remains broader than that of VQF. A contributing mechanism is the iterative active-set update: retaining more features in one round changes the fitted model and can increase the KKT scores of related inactive candidates, allowing additional candidates to enter in subsequent rounds. ADMM-Bisect and WGL($\lambda{=}1$) are therefore useful when broader feature coverage is preferred to highly compact selection.

\paragraph{Support collapse: default WGL.} Under the default continuation schedule, WGL returns an empty support with AUC $=0.500$ on all 10 civil-unrest datasets, whose positive-class fractions are at most $1.52\%$. In contrast, the flu dataset, with a positive-class fraction of $7.4\%$, does not fully collapse. Together with the $\lambda$ sensitivity analysis in Appendix~\ref{app:wgl_sensitivity}, these results indicate that WGL is particularly sensitive to the regularization scale in the severely imbalanced setting. WGL-Adaptive exhibits a related failure pattern, producing very small supports as its adaptive weighting increasingly favors already dominant candidates over weaker ones. These results motivate explicit $\lambda$ calibration before applying WGL to severely imbalanced datasets.

% --------------------------------------------------------------------------
\subsection{Hyperparameter Sensitivity and Robustness}
\label{sec:comparison_hyperparam}

The methods differ substantially in their tuning requirements. VQF introduces no optimization-specific hyperparameters, ADMM-Bisect adds the ADMM penalty $\rho$, and WGL requires careful calibration of the regularization scale and continuation schedule.

\paragraph{VQF: no optimization-specific tuning.} VQF uses only the shared penalty $C$ and active-set threshold $\tau$, which are also required by the other methods and can be selected using validation AUC. It introduces no additional optimization-specific hyperparameters, making it particularly suitable when the available tuning budget is limited.

\paragraph{ADMM-Bisect: limited sensitivity to $\rho$.} ADMM-Bisect additionally requires the augmented-Lagrangian penalty $\rho$. In our sensitivity study, varying $\rho$ over $\{0.1,0.5,1.0,2.0\}$ changes AUC by less than $0.010$ on the LatAm benchmarks. This indicates that, within the tested range, $\rho$ primarily controls optimization behavior while predictive performance remains stable. We therefore use $\rho{=}1.0$ as the default across all 11 benchmarks. ADMM-Bisect provides broader support recovery at the cost of one additional tuning parameter.

\paragraph{WGL: strong sensitivity to $\lambda$.} WGL is considerably more sensitive to its regularization scale. The default continuation schedule $\lambda_{\mathrm{init}}{=}500\to\lambda_{\mathrm{target}}{=}50$ produces empty supports on all 10 severely imbalanced civil-unrest datasets, whereas reducing the regularization scale restores nonzero supports and predictive performance. With $\lambda{=}1$, WGL achieves a mean AUC improvement of $+0.044$ over Logistic Regression ($p{=}0.0032$), which remains significant after Bonferroni correction. The full $\lambda$ sensitivity analysis is reported in Appendix~\ref{app:wgl_sensitivity}. These results show that WGL requires explicit $\lambda$ calibration, particularly on severely imbalanced datasets and when a continuation schedule is used.
% --------------------------------------------------------------------------
\subsection{Optimization Selection in Practice}
\label{sec:comparison_guidelines}

The comparison above provides practical guidance based on the desired balance between compact feature selection, broader support recovery, computational cost, and tuning requirements. When sparse and interpretable selection is important, VQF provides the most compact solution among the tested methods. It selects approximately $4$--$5$ NCFs per dataset, introduces no optimization-specific hyperparameters beyond the shared penalty $C$ and active-set threshold $\tau$, and achieves the largest improvement over Logistic Regression ($\Delta\mathrm{AUC}{=}{+}0.074$, $p{=}4{\times}10^{-5}$), which remains significant after Bonferroni correction. It is therefore well suited to applications where the selected indicators should remain compact and auditable.

When broader feature coverage is desired, ADMM-Bisect provides an alternative with approximately 10 selected NCFs. It improves over Logistic Regression by $\Delta\mathrm{AUC}{=}{+}0.034$ ($p{=}0.0124$), which also passes the Bonferroni-corrected threshold, and is approximately $1.3{\times}$ faster than VQF on the same hardware. ADMM-Bisect introduces one additional hyperparameter, the ADMM penalty $\rho$, while predictive performance varies by less than $0.010$ across $\rho\in\{0.1,0.5,1.0,2.0\}$ in our sensitivity study. We therefore use $\rho{=}1.0$ as the default in the reported experiments.

WGL provides broader feature selection and direct control over the regularization scale, but requires more careful calibration. With $\lambda{=}1$, WGL improves over Logistic Regression by $\Delta\mathrm{AUC}{=}{+}0.044$ ($p{=}0.0032$), which remains significant after Bonferroni correction. In contrast, the default high-to-low continuation schedule produces empty supports on all 10 severely imbalanced civil-unrest datasets. We therefore recommend checking the $\lambda$ sensitivity before applying WGL in highly imbalanced settings rather than relying on the default continuation schedule. The full sensitivity analysis is provided in Appendix~\ref{app:wgl_sensitivity}, and Table~\ref{tab:optimizer_comparison} summarizes the main differences among the optimization methods.
% =============================================================================
% Appendix D: Proposition 4 (Local Heuristics under Ancestor Coupling)
% =============================================================================

\end{document}